\documentclass[nohyperref]{article}

\usepackage{hyperref}

\usepackage[accepted]{icml2022}

\usepackage{times}
\usepackage{epsfig}
\usepackage{graphicx}
\usepackage{amsmath}
\usepackage{amssymb}
\usepackage[british]{babel}
\usepackage{csquotes}
\usepackage{graphicx}
\usepackage{subcaption}
\usepackage{booktabs}
\usepackage{physics}
\usepackage{amsmath}

\usepackage{tabu}
\usepackage{multirow, makecell}
\newcommand\T{\rule{0pt}{2.9ex}}       
\newcommand\B{\rule[-1.2ex]{0pt}{0pt}} 

\usepackage[capitalize,noabbrev]{cleveref}

\usepackage[textsize=tiny]{todonotes}

\icmltitlerunning{Skin Deep Unlearning}

\begin{document}

\twocolumn[
\icmltitle{Skin Deep Unlearning: Artefact and Instrument Debiasing in the Context of Melanoma Classification}




\begin{icmlauthorlist}
\icmlauthor{Peter J. Bevan}{newc}
\icmlauthor{Amir Atapour-Abarghouei}{dur}
\end{icmlauthorlist}

\icmlaffiliation{newc}{School of Computing, Newcastle University, Newcastle upon Tyne, UK}
\icmlaffiliation{dur}{Department of Computer Science, Durham University, Durham, UK}

\icmlcorrespondingauthor{Peter Bevan}{peterbevan@hotmail.co.uk}

\icmlkeywords{Machine Learning, ICML, Computer Vision, CNN, Debiasing, Melanoma}

\vskip 0.3in
]



\printAffiliationsAndNotice{}  

\begin{abstract}
Convolutional Neural Networks have demonstrated dermatologist-level performance in the classification of melanoma from skin lesion images, but prediction irregularities due to biases seen within the training data are an issue that should be addressed before widespread deployment is possible. In this work, we robustly remove bias and spurious variation from an automated melanoma classification pipeline using two leading bias \enquote{unlearning} techniques. We show that the biases introduced by surgical markings and rulers presented in previous studies can be reasonably mitigated using these bias removal methods. We also demonstrate the generalisation benefits of \enquote{unlearning} spurious variation relating to the imaging instrument used to capture lesion images. Our experimental results provide evidence that the effects of each of the aforementioned biases are notably reduced, with different debiasing techniques excelling at different tasks.
\end{abstract}

\section{Introduction}
In recent years, Convolutional Neural Networks (CNN) have demonstrated skin lesion diagnosis performance on par with experienced dermatologists \cite{brinkerConvolutionalNeuralNetwork2019,brinkerComparingArtificialIntelligence2019,haenssleManMachineDiagnostic2018a}. This is particularly important since, when diagnosed early, melanoma may be easily cured by surgical excision \cite{haenssleManMachineReloaded2020,winklerAssociationSurgicalSkin2019}, and so accessible and accurate diagnostic tools have the potential to democratise dermatology and save numerous lives worldwide.

\begin{figure}[t]
\includegraphics[width=\columnwidth]{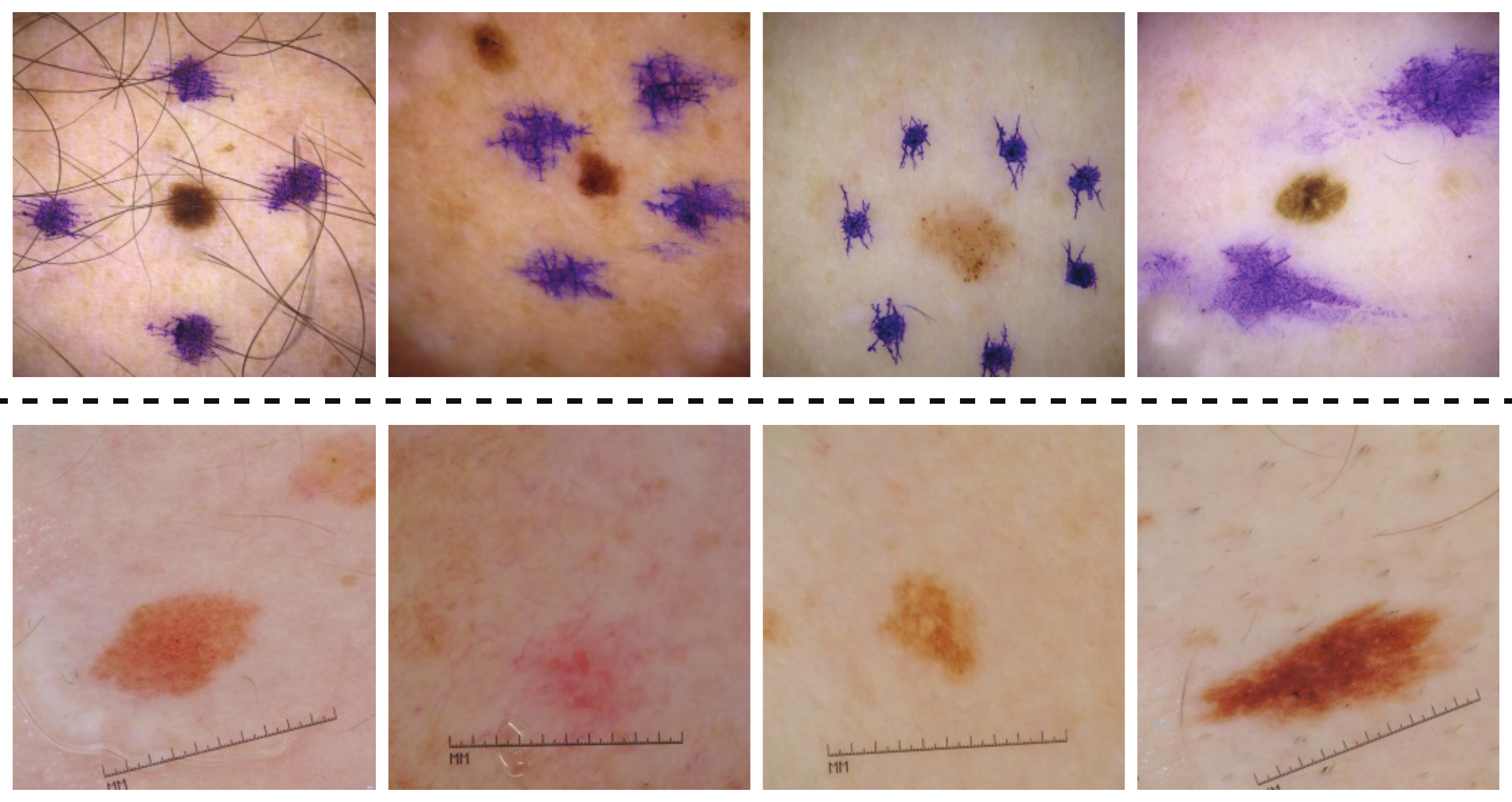}
\centering
  \caption{Examples of artefacts seen in ISIC 2020 data \cite{rotembergPatientcentricDatasetImages2021}. Top row shows images with surgical markings present, bottom row shows images with rulers.}
\label{fig:Lesions}
\end{figure}

While deploying such learning-based techniques far and wide could be massively beneficial, great care must be taken as any small pitfall could be replicated on a massive scale. For example, some dermatologists use visual aids such as skin markings to mark the location of a lesion, or rulers to indicate scale, as seen in \Cref{fig:Lesions}. In fact, Winkler et al. \cite{winklerAssociationDifferentScale2021,winklerAssociationSurgicalSkin2019} demonstrated how bias induced by the presence of these artefacts can result in diminished classification performance. They also suggest that dermatologists avoid using these aids in the future, which is a valid solution to the problem, though changing the habits of every dermatologist is highly unrealistic and could potentially be detrimental to their performance.

Segmentation of the lesion from the surrounding skin has also previously been proposed, but is not a good option, since \enquote*{any kind of pre-processing or segmentation itself
may erroneously introduce changes that impede a CNN’s
correct classification of a lesion} \cite{winklerAssociationDifferentScale2021}. Cropping surgical markings out of the image has been shown to be effective at mitigating surgical marking bias in the work of \citet{winklerAssociationSurgicalSkin2019}, but it is noted this must be done by an experienced dermatologist to prevent the loss of important information, which is costly and time-consuming.

Consequently, the alternative path towards diminishing the effects of such artefacts would be not to remove the artefacts themselves from the image, but to reduce their influence on how the model functions, which translates to removing the \enquote{bias} these artefacts introduce into the learning process. As such, recent advances in debiasing architectures for CNNs \cite{alviTurningBlindEye2019,kimLearningNotLearn2019a} present an excellent opportunity to robustly mitigate the aforementioned biases without any need to alter the behaviour of physicians or pre-process the image data.

Surgical artefacts left by physicians are not the only concern when it comes to skin lesion classification, however. Another issue that plagues many machine learning models is the domain shift between the training and real-world inference data, leading models to perform poorly upon deployment. One cause of this domain shift in skin lesion classification is likely to be spurious variation from minor differences in the imaging instruments used to capture lesions. Inspired by the work of \citet{ganinDomainAdversarialTrainingNeural2017}, we propose also using \enquote{unlearning} techniques \cite{alviTurningBlindEye2019, kimLearningNotLearn2019a} for domain generalisation by removing spurious variation associated with instrument type to create a more generalisable, instrument-invariant model.

In summary, this work aims to explore bias and domain \enquote{unlearning} towards creating more robust, generalisable and fair models for melanoma classification. All implementation is publicly available on \href{https://github.com/pbevan1/Skin-Deep-Unlearning}{GitHub}\footnote{\url{https://github.com/pbevan1/Skin-Deep-Unlearning}}. Our primary contributions can be summarised as follows:
\begin{itemize}
  \item \textit{Melanoma classification} - The models presented demonstrate impressive melanoma classification performance, with many of the tested architectures beating the average performance of experienced dermatologists on a benchmark dataset\footnote{It should be noted that this is based on the classification of a single lesion image and in practice dermatologists have access to further sources of information during patient consultation.} \cite{brinkerComparingArtificialIntelligence2019} (\Cref{subsec:experiments:results:domgen}).
  \item \textit{Artefact debiasing} - We mitigate the bias introduced by surgical markings and rulers, as shown in the work of \citet{winklerAssociationDifferentScale2021,winklerAssociationSurgicalSkin2019} using \enquote{Learning Not to Learn} \cite{kimLearningNotLearn2019a} and \enquote{Turning a Blind Eye} \cite{alviTurningBlindEye2019} methods (\Cref{subsec:experiments:results:artefacts}).
  \item \textit{Domain generalisation} - We demonstrate the generalisation benefits of unlearning \cite{alviTurningBlindEye2019,kimLearningNotLearn2019a} information relating to the instruments used to capture skin lesion images (\Cref{subsec:experiments:results:domgen}).
\end{itemize}
We introduce related work in \Cref{sec:related}, methods in \Cref{sec:methods}, experimental results in \Cref{sec:results}, limitations and potential future work in \Cref{sec:limitations}, and conclusions in \Cref{sec:conclusions}.

\section{Related work}
\label{sec:related}

We consider related work within two distinct areas, namely artefact bias in skin lesion images (\Cref{sec:related:artefact}) and domain generalisation (\Cref{sec:related:domain}).

\subsection{Artefacts bias}
\label{sec:related:artefact}
One of the problems addressed in this paper is the algorithmic bias introduced by certain artefacts present in skin lesion images. Precedent for investigating the mitigation of artefact bias in skin lesion classification is found in the work of \citet{winklerAssociationSurgicalSkin2019}, which compares the performance of a CNN classification model on 130 lesions \textit{without} surgical markings present, versus the same 130 lesions \textit{with} surgical markings present. Strong bias was demonstrated, with specificity hit hard, as well as Area Under the Curve (AUC).
Another work \cite{winklerAssociationDifferentScale2021} shows a similar level of bias caused by rulers in skin lesion images.

Segmentation of skin lesions from the surrounding healthy skin has been suggested as a means of removing artefacts from the input of skin lesion classification models \cite{jafariSkinLesionSegmentation2016,okurSurveyAutomatedMelanoma2018a}. However, this is not commonly used at the time of writing, given that CNNs may utilise information in the surrounding skin regions  \cite{bissotoConstructingBiasSkin2019b,winklerAssociationDifferentScale2021} and so removing this can impact classification performance. Artefacts themselves also often impede segmentation \cite{mishraOverviewMelanomaDetection2016}, and artefacts that are on the lesion itself are not separable from the lesion by image region.

In an earlier work, \citet{bissotoDebiasingSkinLesion2020} tackle the issue of artefact bias removal in a manner similar to the one proposed in this work by using a model with seven debiasing heads in an attempt to mitigate the bias caused by seven artefacts. The authors conclude that the bias removal method in the work of \citet{kimLearningNotLearn2019a} (\enquote{Learning Not to Learn}) is not ready to tackle the issue. However, ablation studies to isolate each head are lacking, and so the efficacy of each of the seven individual debiasing heads cannot be ascertained. It is, therefore, possible that certain heads bring down the performance of the entire model, or interact with each other unfavourably. In addition to this, the paper does not experiment with other leading debiasing solutions such as the one proposed in the work of \citet{alviTurningBlindEye2019} (\enquote{Turning a Blind Eye}), which may be more effective at the given task.

In this work, we only focus on  biases that are well documented as causing performance degradation, and compare individual debiasing heads across different methods before combining these heads. \citet{bissotoDebiasingSkinLesion2020} do note improvements in performance when testing their debiasing models on data with significant domain shift such as the Interactive Atlas of Dermoscopy clinical data \cite{lioInteractiveAtlasDermoscopy2004}, which indicates some improvement in generalisation. We build upon this notion in our domain generalisation experiments (\Cref{subsec:experiments:results:domgen}).

\subsection{Domain generalisation}
\label{sec:related:domain}

\begin{figure}[t]
\includegraphics[width=0.9\columnwidth]{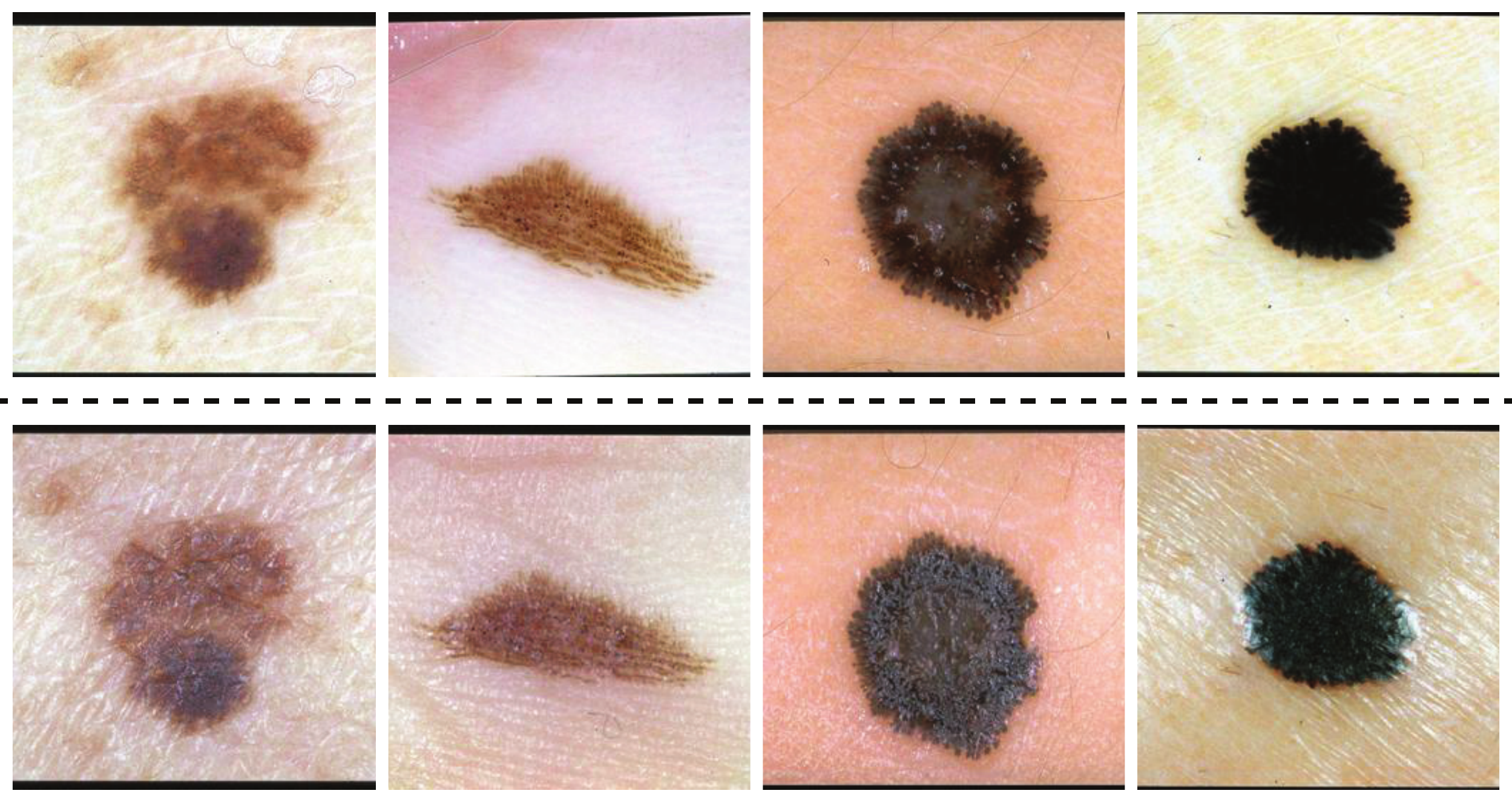}
\centering
  \caption{Domain shift between clinical and dermoscopic images of the same lesion \cite{lioInteractiveAtlasDermoscopy2004}. Top row shows dermoscopic images, bottom row clinical.}
\label{fig:clinical-derm}
\end{figure}

A common assumption in machine learning is that the training and test data are drawn from the same distribution, though this assumption does not usually hold true in real-world applications \cite{csurkaComprehensiveSurveyDomain2017a}. For instance, inconsistencies in prostate cancer classification performance between image samples originating from different clinics is shown by \citet{arvidssonGeneralizationProstateCancer2018}, and the authors hypothesise that this could be due to domain shift caused by variation in the equipment used. In skin lesion classification, there are two main imaging methods: dermoscopic (skin surface microscopy), and clinical (standard photograph) \cite{westerhoffIncreaseSensitivityMelanoma2000a} (see \Cref{fig:clinical-derm}). This domain shift has been shown to impact model performance \cite{guProgressiveTransferLearning2020}. Additionally, within these two imaging methods, many different brands and models of instrument are used by different clinics, which may also introduce domain bias. Supporting this hypothesis, it is shown by \citet{jacksonCameraBiasFine2020} that CNNs can easily discriminate between camera models, which can lead models to overfit to this spurious variation during training.

Domain adaptation methods have been successfully used to minimise the distance between the underlying distributions of training and test datasets, i.e. a model trained on a given dataset (source distribution) is enabled to perform well on a different dataset (target distribution) via domain adaptation \cite{atapour-abarghoueiRealTimeMonocularDepth2018,csurkaComprehensiveSurveyDomain2017a,guProgressiveTransferLearning2020,guanDomainAdaptationMedical2021}. However, such methods require knowledge of the target distribution, which is not always readily available. Domain generalisation, on the other hand, is more robust than domain adaptation, and differs in that the target domain is unseen \cite{liDomainGeneralizationMedical2020}, aiming for improved performance on a wide range of possible test data. In this work, we explore applying bias unlearning techniques \cite{alviTurningBlindEye2019,kimLearningNotLearn2019a} towards domain generalisation in melanoma classification, attempting to find an instrument-invariant feature representation without compromising performance.

\section{Methods}
\label{sec:methods}

In this work, two leading debiasing techniques within the literature are used, namely \enquote{Learning Not To Learn} (LNTL) \cite{kimLearningNotLearn2019a} and \enquote{Turning a Blind Eye} (TABE) \cite{alviTurningBlindEye2019}. Both of these are often referred to as \enquote{unlearning} techniques because of their ability to remove bias from the feature representation of a network by minimising the mutual information between the feature embedding and the unwanted bias. Further details of these unlearning methods are described in \Cref{subsec:methods:lntl,subsec:methods:tabe}.

\subsection{Learning Not to Learn}
\label{subsec:methods:lntl}

\begin{figure}[htbp]
\includegraphics[width=\columnwidth]{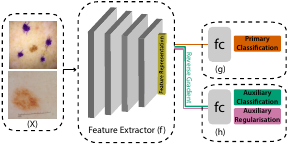}
\centering
\caption{\enquote{Learning Not to Learn} architecture. Feature extractor, $f$, is implemented as a convolutional architecture such as ResNeXt or EfficientNet in this work. \enquote{fc} denotes a fully-connected layer.}
\label{fig:LNTL}
\end{figure}

\enquote{Learning Not to Learn} (LNTL) \cite{kimLearningNotLearn2019a} proposes a novel regularisation loss combined with a gradient reversal layer \cite{ganinDomainAdversarialTrainingNeural2017} to remove bias from the feature representation of a CNN during backpropagation. \Cref{fig:LNTL} shows a generic overview of the LNTL architecture. The input image, $x$, is passed into a feature extractor, $f$: $x \rightarrow \mathbb{R}^K$, where $K$ is the dimension of the embedded feature.

The feature extractor is implemented as a pre-trained convolutional architecture such as ResNeXt \cite{heDeepResidualLearning2016} or EfficientNet \cite{tanEfficientNetRethinkingModel2019} in this work. The extracted feature embedding is then passed in parallel into both $g$: \(\mathbb{R}^K \rightarrow \mathcal{Y}\) and $h$: \(\mathbb{R}^K \rightarrow \mathcal{B}\), the primary and auxiliary classification heads respectively, where, in the case of this work, \(\mathcal{Y}\) represents the set of possible lesion classes and \(\mathcal{B}\) represents the set of target bias classes.

The networks $f$ and $h$ play the minimax game, in which $h$ is trained to classify the bias from the extracted feature embedding (minimising cross-entropy), whilst $f$ is trained to maximise the cross-entropy to restrain $h$ from predicting the bias, and also to minimise the negative conditional entropy to reduce the mutual information between the feature representation and the bias. The gradient reversal layer between $h$ and $f$ acts as an additional step to remove information relating to the target bias from the feature representation.

The gradient reversal layer works by multiplying the gradient of the auxiliary classification loss by a negative scalar during backpropagation, causing the feature extraction network, $f$, to \enquote{learn not to learn} the targeted bias, $b(x)$, rather than learn it. By the end of training, $f$ has learnt to extract a feature embedding independent of the bias, $g$ has learnt to use this feature embedding to perform the primary classification task without relying on the bias, and $h$ performs poorly at predicting the bias due to the lack of bias information in the feature embedding. 

The minimax game along with the main classification loss are formulated as:

\begin{equation}
	\begin{split} 
		\min\limits_{\theta_f,\theta{g}} \max\limits_{\theta_h} \ & \mathbb{E}_{\tilde{x}\sim P_X(\cdot)}  [\underbrace{\mathcal{L}_g(\theta_f,\theta_g)}_{(a)}\\
		& + \underbrace{\lambda\mathbb{E}_{\tilde{b}\sim Q(\cdot|f(\tilde{x}))} [\log{Q}(\tilde{b}|f(\tilde{x}))]]}_{(b)}\\
		& - \underbrace{\mu\mathcal{L}_\mathcal{B}(\theta_f,\theta_h)}_{(c)},
		\label{eq:LNTL}
	\end{split}
\end{equation}
where (a) represents the cross-entropy loss of the main classification head, (b) represents the regularisation loss and (c) represents the cross-entropy loss of the auxiliary bias classification head. The hyperparameters $\lambda$ and $\mu$ are used to balance the terms. The parameters of each network are denoted as $\theta_f$, $\theta_g$ and $\theta_h$. An auxiliary distribution, $Q$, is used to approximate the posterior distribution of the bias, $P$, which is paramaterised as the bias prediction network, $h$.

\subsection{Turning a Blind Eye}
\label{subsec:methods:tabe}

\begin{figure}[htbp]
\includegraphics[width=\columnwidth]{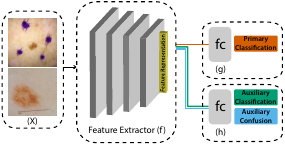}
\centering
\caption{\enquote{Turning a Blind Eye} generic architecture. Feature extractor, $f$, is implemented as a convolutional architecture such as ResNeXt or EfficientNet in this work. \enquote{fc} denotes a fully-connected layer.}
\label{fig:TABE}
\end{figure}

\Cref{fig:TABE} shows a generic overview of the \enquote{Turning a Blind Eye} (TABE) \cite{alviTurningBlindEye2019} architecture. Similar to LNTL \cite{kimLearningNotLearn2019a}, this method also removes unwanted bias using an auxiliary classifier, $\theta_m$, where $m$ is the $m$-th unwanted bias. The TABE auxiliary classifier minimises an auxiliary classification loss, $\mathcal{L}_s$, used to identify bias in the feature representation, $\theta\textsubscript{repr}$, as well as an auxiliary confusion loss \cite{tzengSimultaneousDeepTransfer2015a}, $\mathcal{L}\textsubscript{conf}$, used to make $\theta\textsubscript{repr}$ invariant to the unwanted bias. Since these losses stand in opposition to one another, they are minimised in separate steps: first $\mathcal{L}_s$ alone, and then the primary classification loss, $\mathcal{L}_p$, together with $\mathcal{L}\textsubscript{conf}$. The confusion loss is defined as follows:
\begin{equation} \label{eq:TABE1}
\mathcal{L}\textsubscript{conf,m}(x_m,y_m,\theta_m;\theta\textsubscript{repr}) = -\\\sum_{n_m}\frac{1}{n_m}{\log}{p_n}_m,
\end{equation}
where $x_m$ is the input, $y_m$ is the bias label, ${p_n}_m$ is the softmax of the auxiliary classifier output and $n_m$ is the number of auxiliary classes. This confusion loss works towards finding a representation in which the auxiliary classification head performs poorly by finding the cross entropy between the output predicted bias and a uniform distribution. The complete joint loss function being minimised is:
\begin{equation}
	\begin{split} 
		\mathcal{L}(x_p,y_p,x_s,y_s,\theta_p,\theta_s,\theta\textsubscript{repr}) & = \mathcal{L}_p(x_p,y_p;\theta\textsubscript{repr},\theta_p)\\
		& + \mathcal{L}_s+\alpha\mathcal{L}\textsubscript{conf},
		\label{eq:TABE2}
	\end{split}
\end{equation}
where $\alpha$ is a hyperparameter which determines how strongly the confusion loss impacts the overall loss. The feature extractor, $f$, is implemented as a pre-trained convolutional architecture such as ResNeXt \cite{heDeepResidualLearning2016} or EfficientNet \cite{tanEfficientNetRethinkingModel2019} in this work.

Additionally, as proposed by \citet{kimLearningNotLearn2019a}, a hybrid of LNTL and TABE can be created by utilising the confusion loss (CL) from TABE, and then also applying gradient reversal (GR) from LNTL to the auxiliary classification loss as it is backpropegated to $f$. We denote this configuration as \enquote{CLGR}.

\subsection{Datasets}
This section briefly describes the datasets used in the experiments (see supplementary material Section \ref{appendix:Examples} for example images).

\subsubsection{ISIC challenge training data}
\label{subsubsec:ISIC_data}
The International Skin Imaging Collabaration (ISIC) challenge is a yearly automated melanoma classification challenge with several publicly available dermoscopic skin lesion datasets (see \href{https://www.isic-archive.com/}{ISIC archive}\footnote{\url{https://www.isic-archive.com/}}), complete with diagnosis labels and metadata. A combination of the 2017 and 2020 ISIC challenge data \cite{codellaSkinLesionAnalysis2018,rotembergPatientcentricDatasetImages2021} (35,574 images) is used as the training data in this work due to the higher representation of artefacts in these datasets than other competition years. Pre-processed (centre cropped and resized) images of size 256$\times$256 are used for all training and testing. The surgical markings are labelled using colour thresholding, with the labels double-checked manually, while the rulers are labelled entirely manually. A random subset (33\%, 3,326 images) of the 2018 \cite{codellaSkinLesionAnalysis2018} challenge data is used as the validation set for hyperparameter tuning.

The model and training data used by \citet{winklerAssociationDifferentScale2021,winklerAssociationSurgicalSkin2019} are proprietary, and so the bias in these studies could not be exactly reproduced. Alternatively, since the primary objective is to investigate the possibility of removing bias from the task, we skew the ISIC data \cite{codellaSkinLesionAnalysis2018,rotembergPatientcentricDatasetImages2021} to produce similar levels of bias in our baseline model as shown in the aforementioned studies. Benign lesions in the training data with surgical markings are removed and images that are both malignant and marked are duplicated and randomly augmented (treating each duplicate as a new data point) to skew the model towards producing false positives for lesions with surgical markings, thus reproducing the level of bias demonstrated by \citet{winklerAssociationSurgicalSkin2019}. The dataset is processed similarly with rulers to demonstrate ruler bias. The number of duplications of melanoma images with surgical markings present, $dm$, and with rulers present, $dr$, are used as hyperparameters to control the level of skew in experiments. This artificially skewed data is only used to demonstrate artefact debiasing (\Cref{subsec:experiments:results:artefacts}), and the original data is used for all other experiments.

\subsubsection{Heidelberg University test data}
The test set presented by \citet{winklerAssociationSurgicalSkin2019} is used to evaluate the artefact debiasing approach presented in this work (\Cref{subsec:experiments:results:artefacts}). The dataset comprises 130 lesions: 23 malignant, 107 benign. There are two images of each lesion in the set, one without surgical markings, and one with surgical markings. This allows a direct evaluation of the effect of surgical marking bias on the performance of a model. The test set from the ruler bias study \cite{winklerAssociationDifferentScale2021} is not publicly available or shared, so the plain images from the work of \citet{winklerAssociationSurgicalSkin2019} are superimposed with rulers to be used as test images. The approach of superimposing rulers was validated as not statistically significantly different from in-vivo rulers by \citet{winklerAssociationDifferentScale2021}.

\subsubsection{MClass benchmark test data}
The MClass public human benchmark introduced by \citet{brinkerComparingArtificialIntelligence2019} is used as a test dataset for assessing domain generalisation (\Cref{subsec:experiments:results:domgen}), also providing a human benchmark. This dataset comprises a set of 100 dermoscopic images and 100 clinical images (\textit{different} lesions), each with 20 malignant and 80 benign lesions. In the study, the dermoscopic and clinical image sets were classified by 157 and 145 experienced dermatologists respectively, with their average classification performances published. The dermoscopic MClass data is made up of images from the ISIC archive, some of which were also present in our ISIC training data (\Cref{subsubsec:ISIC_data}), so these were removed from the training data to prevent data leakage.

\subsubsection{Interactive Atlas of Dermoscopy and Asan data}
Two additional test sets, the Interactive Atlas of Dermoscopy dataset \cite{lioInteractiveAtlasDermoscopy2004}, and the Asan test dataset \cite{hanClassificationClinicalImages2018}, are used to further test domain generalisation (\Cref{subsec:experiments:results:domgen}). The Atlas dataset comprises 1,011 lesions across 7 classes, with one dermoscopic and one clinical image per lesion. The Asan test dataset comprises 852 clinical images across 7 classes of lesions. Whilst the ISIC training data \cite{codellaSkinLesionAnalysis2018,rotembergPatientcentricDatasetImages2021} is mostly white Western patients, the Atlas and Asan datasets have representation from a broader variety of ethnic groups, which enables a better test of a model's ability to deal with domain shift.

\subsection{Implementation}

All experiments are implemented in PyTorch \cite{NEURIPS2019_9015} and carried out using two NVIDIA Titan RTX GPUs in parallel with a combined memory of 48 GB on an Arch Linux system with a 3.30GHz 10-core Intel CPU and 64 GB of memory. The baseline model is inspired by the winning entry from the 2020 ISIC challenge \cite{haIdentifyingMelanomaImages2020a}, which utilises the EfficientNet-B3 architecture \cite{tanEfficientNetRethinkingModel2019}, pre-trained on the ImageNet dataset \cite{dengImageNetLargescaleHierarchical2009}. ResNet-101 \cite{heDeepResidualLearning2016}, ResNeXt-101 \cite{xieAggregatedResidualTransformations2017}, DenseNet \cite{huangDenselyConnectedConvolutional2017} and Inception-v3 \cite{szegedyRethinkingInceptionArchitecture2016a} are each substituted for EfficientNet-B3 to evaluate the optimal network for the task, simultaneously testing the effectiveness of the debiasing techniques across different architectures.

Early experimentation showed ResNeXt-101 to be the overall best performing architecture, as seen in \Cref{tab:Instrument}, and therefore experimentation is focused around using this as the feature extractor in the domain generalisation experiments. EfficientNet-B3 is kept as the base architecture for surgical marking and ruler debiasing since the baseline performance is closest to the reported performance of the unknown proprietary model used in the work of \citet{winklerAssociationSurgicalSkin2019}. The primary and auxiliary classification heads are implemented as a single fully-connected layer, as suggested by \citet{kimLearningNotLearn2019a}.
Stochastic gradient descent is used across all models, ensuring comparability and compatibility between the baseline and debiasing networks.

Following a grid search, the learning rate (searched between 0.03 and 0.00001) and momentum (searched between 0 and 0.9) are selected as 0.0003 and 0.9 respectively (see Section \ref{appendix:Val} of the supplementary material for full hyperparameter tuning results). The learning rate of the TABE heads is boosted by a factor of 10 (to 0.003), as suggested by \citet{alviTurningBlindEye2019}, except when using multiple debiasing heads since this seems to cause instability. The best performing values of the hyperparameters $\alpha$ and $\lambda$ in \Cref{eq:LNTL} and \Cref{eq:TABE2} are also empirically chosen to be $\alpha=0.03$ and $\lambda=0.01$.

A weighted loss function is implemented for all auxiliary heads to tackle class imbalance, with each weighting coefficient, $\mathcal{W}_n$, being the inverse of the corresponding class frequency, $c$.
Since the proportion of benign and malignant lesions is highly imbalanced in the test sets, accuracy proved not to be a descriptive metric to use. Instead, AUC is used as the primary metric across all experiments, as is standard in melanoma classification \cite{haIdentifyingMelanomaImages2020a,hanClassificationClinicalImages2018,liSkinLesionAnalysis2018,okurSurveyAutomatedMelanoma2018a}, given that it takes into account both sensitivity and specificity across all thresholds and is effective at communicating the performance when the target classes are imbalanced \cite{mandrekarReceiverOperatingCharacteristic2010}. We use test-time augmentation \cite{Simonyan15,szegedyRethinkingInceptionArchitecture2016a} to average predictions over 8 random flips along different axes, applied to all test images, to enable a fairer evaluation of our models. The optimal number of epochs for training each architecture on each dataset is chosen through analysis of the 5-fold cross validation curves for the baseline models, selecting the epoch at which the AUC reached its maximum or plateaued (see Section \ref{tab:Epochs} of the supplementary material).

\section{Experimental results}
\label{sec:results}

The results of our artefact bias removal experiments are presented in \Cref{subsec:experiments:results:artefacts}. We present the domain generalisation experimental results in \Cref{subsec:experiments:results:domgen}. The three variations of debiasing heads that were implemented and combined in the experiments are: \enquote{Learning Not to Learn} (\textbf{LNTL}) \cite{kimLearningNotLearn2019a}, \enquote{Turning a Blind Eye} (\textbf{TABE}) \cite{alviTurningBlindEye2019}, and a hybrid of the confusion loss (CL) from TABE with the gradient reversal layer (GRL) from LNTL (\textbf{CLGR}).

\subsection{Artefacts bias removal}
\label{subsec:experiments:results:artefacts}
\begin{figure}[htbp]
\includegraphics[width=\columnwidth]{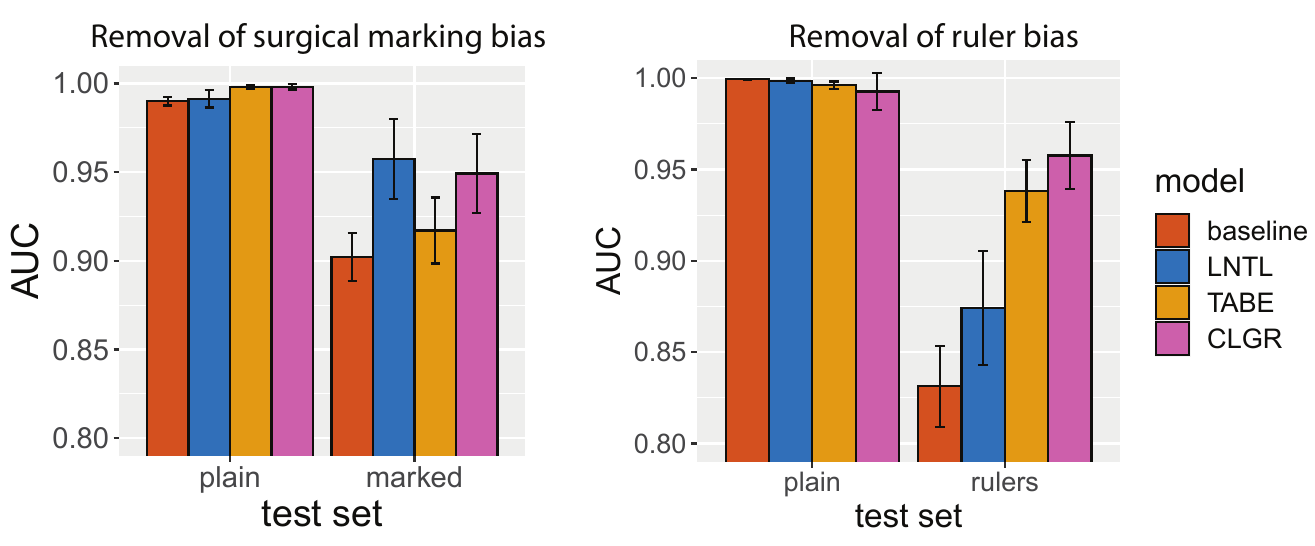}
\centering
\caption{Comparing melanoma classification performance of artefact debiasing models against the baseline, trained on artificially skewed ISIC data.}
\label{fig:RulerMarked}
\end{figure}

We attempt to mitigate the bias caused by two artefacts that have been shown to affect performance in melanoma classification, namely surgical markings \cite{winklerAssociationSurgicalSkin2019} and rulers \cite{winklerAssociationDifferentScale2021} (see \Cref{tab:RulerMarked} and \Cref{fig:RulerMarked}). Separate individually-skewed training sets are used with skew levels set at $dm$=20 (dm being number of duplications of images with surgical markings) for examining the removal of surgical marking bias and $dr$=18 (dr being duplications of images with rulers) for ruler bias. We use surgical marking and ruler labels as the target for the debiasing heads in each of these experiments respectively.

Each model is trained and evaluated 6 times using 6 different random seeds, allowing the mean and standard deviation to be reported. The high scores are due to the inherent clarity of the cues within the images and consequent simplicity of the classification of the test set, consistent with the scores reported by \citet{winklerAssociationDifferentScale2021,winklerAssociationSurgicalSkin2019}. Any chance of a leak between the test set and the ISIC training data has been ruled out \footnote{The corresponding author of the test set was contacted and they ruled out a data leak between their test set and the ISIC data.}. Despite the ease of classification, both the existence of bias and the effectiveness of bias mitigation can still be demonstrated, and experiments using the original test set provide direct evidence that we are able to mitigate the problem presented in these studies. While the baseline model performs very well for the unbiased images (\Cref{tab:RulerMarked} - \enquote{Heid Plain}), performance suffers when this model is tested on the same lesions with either artefact present, replicating the findings from the work of \citet{winklerAssociationDifferentScale2021,winklerAssociationSurgicalSkin2019}.

\begin{table*}[!t]
\captionsetup[table]{skip=7pt}
\captionof{table}{\textit{Artefact debiasing}: Melanoma classification performance when using unlearning techniques versus a baseline, trained on skewed ISIC data. Scores are \textbf{AUC}. \enquote{Heid Plain} is free from artefacts while \enquote{Heid Marked} and \enquote{Heid Rulers} contain the same lesions with surgical markings and rulers present. $\dagger$ indicates surgical marking labels as target for the auxiliary head and $\ddagger$ indicates ruler labels.}
	\centering
	\resizebox{0.83\linewidth}{!}{
		{\tabulinesep=0mm
			\begin{tabu}{@{\extracolsep{12pt}}c c c@{}}
				\hline\hline
				\multicolumn{1}{c}{\multirow{2}{*}{Experiment}} & 
				\multicolumn{2}{c}{(a) Surgical Marking Removal ($dm$=20)}\T\B \\
				\cline{2-3}
						& Heid Plain & Heid Marked\T\B\\
				\hline\hline
Baseline & 0.990$\pm$0.002 & 0.902$\pm$0.013\T\\
LNTL$\dagger$ & 0.991$\pm$0.005 & \textbf{0.957}$\pm$0.023\\
TABE$\dagger$ & \textbf{0.998}$\pm$0.001 & 0.917$\pm$0.019\\
CLGR$\dagger$ & 0.998$\pm$0.002 & 0.949$\pm$0.022\B\\

\hline
\hline
        \end{tabu}
        \quad
			\begin{tabu}{@{\extracolsep{12pt}}c c c@{}}
				\hline\hline
				\multicolumn{1}{c}{\multirow{2}{*}{Experiment}} &
				\multicolumn{2}{c}{(b) Ruler Bias Removal ($dr$=18)}\T\B \\
				\cline{2-3}
						& Heid Plain & Heid Ruler\T\B\\
				\hline\hline
Baseline & \textbf{0.999}$\pm$0.000 & 0.831$\pm$0.022\T\\
LNTL$\ddagger$ & 0.997$\pm$0.001 & 0.874$\pm$0.031\\
TABE$\ddagger$ & 0.992$\pm$0.002 & 0.938$\pm$0.017\\
CLGR$\ddagger$ & 0.999$\pm$0.010 & \textbf{0.958}$\pm$0.018\B\\

\hline
\hline
        \end{tabu}
    }
}
\label{tab:RulerMarked}
\end{table*}

\Cref{fig:RulerMarked} presents evidence that each debiasing method is successful at mitigating artefact bias. LNTL is the most effective at unlearning surgical marking bias, improving on the baseline AUC by 0.055 (6.1\% increase) on the equivalent marked images from the same set (\enquote{Heid marked}). All three techniques also mitigate ruler bias well, with CLGR being the most effective and showing a 0.127 increase in AUC compared to the baseline (15.3\% increase). The results of our experiments suggest that unlearning techniques can be used to reduce the bias demonstrated by \citet{winklerAssociationDifferentScale2021,winklerAssociationSurgicalSkin2019}, but are not a perfect solution, given that the artefacts still have a negative impact on performance when compared to performance on the plain images.

\subsection{Domain generalisation}
\label{subsec:experiments:results:domgen}
\begin{figure}[htbp]
\includegraphics[width=\columnwidth]{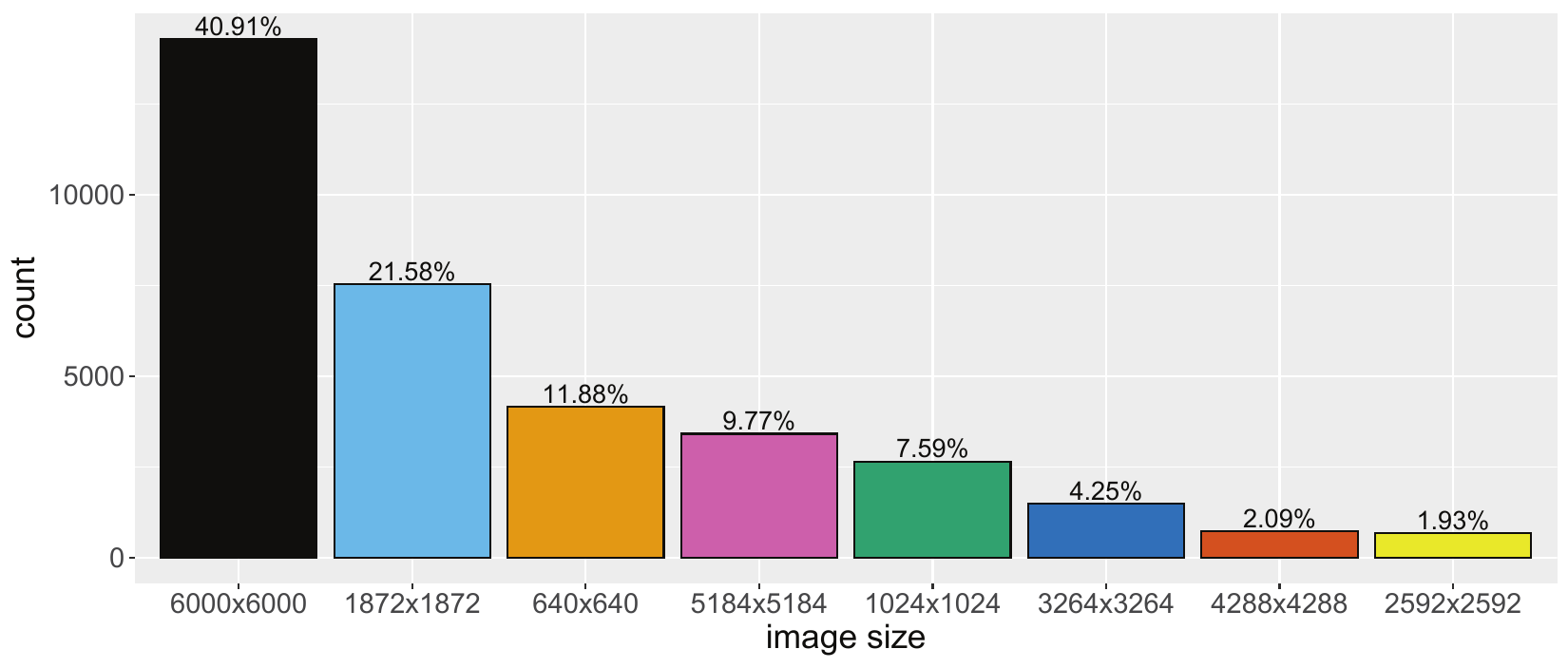}
\centering
\caption{Class distribution of instruments in ISIC 2020/2017 combined data \cite{codellaSkinLesionAnalysis2018,rotembergPatientcentricDatasetImages2021}. Instruments inferred as separate through image size.}
\label{fig:Instrument}
\end{figure}

Another significant issue within melanoma classification is instrument bias, hindering the application of a trained model to image data acquired via different imaging instruments. We attempt to address this issue by removing instrument bias from the model pipeline using unlearning techniques \cite{alviTurningBlindEye2019,kimLearningNotLearn2019a}, with the aim of improving domain generalisation. According to ISIC, image dimensions in ISIC competition data are a good proxy for the imaging instrument used to capture the image\footnote{ISIC were contacted about origin clinics for their images and they mentioned the link between image dimensions and origin.}. \Cref{fig:Instrument} shows the distribution of image sizes in the ISIC dataset, following omission of outlier classes.

These dimensions were used as the auxiliary target for debiasing, attempting to remove spurious variation related to the imaging instrument from the feature representation. The vast majority (98\%) of the ISIC training images \cite{codellaSkinLesionAnalysis2018,rotembergPatientcentricDatasetImages2021} make up the first 8 \enquote{instrument} categories, but there are many outlier categories with a very small number of observations, which are discarded to prevent significant class imbalance.

\begin{table*}[!t]
\captionsetup[table]{skip=7pt}
\captionof{table}{\textit{Domain generalisation}: Comparing generalisation ability of each debiasing method across different architectures, trained using ISIC 2017 and 2020 data. All scores are \textbf{AUC}. The \enquote{dermatologists} row is the AUC scores from the work of \citet{brinkerComparingArtificialIntelligence2019}. The $\mathsection$ symbol indicates the use of instrument labels for the auxiliary head. Bold numbers are the highest score for that architecture, underlined scores are the highest scores across all architectures.}
	\centering
	\resizebox{0.83\linewidth}{!}{
		{\tabulinesep=0mm
			\begin{tabu}{@{\extracolsep{12pt}}c c c c c c c@{}}
				\hline\hline
				\multicolumn{1}{c}{\multirow{2}{*}{Experiment}} & 
				\multicolumn{1}{c}{\multirow{2}{*}{Architecture}} & 
				\multicolumn{2}{c}{Atlas} &
				\multicolumn{1}{c}{Asan} &
				\multicolumn{2}{c}{MClass}\T\B \\
				\cline{3-4} \cline{5-5} \cline{6-7}
						& & Dermoscopic & Clinical & Clinical & Dermoscopic & Clinical\T\B\\
				\hline\hline
Dermatologists & --- & --- & --- & --- & 0.671 & 0.769\T\B\\
\hline

Baseline & EfficientNet-B3 & 0.757 & 0.565 & 0.477 & 0.786 & 0.775\T\\

LNTL$\mathsection$ & EfficientNet-B3 & 0.709 & 0.562 & 0.570 & 0.830 & 0.630 \\

TABE$\mathsection$ & EfficientNet-B3 & \textbf{0.811} & \textbf{0.629} & \textbf{0.685} & 0.877 & \underline{\textbf{0.889}} \\

CLGR$\mathsection$ & EfficientNet-B3 & 0.761 & 0.562 & 0.656 & \textbf{0.882} & 0.838\B\\
\hline
Baseline & ResNet-101 & \textbf{0.802} & 0.606 & 0.704 & \textbf{0.877} & \textbf{0.819}\T\\
LNTL$\mathsection$ & ResNet-101 & 0.776 & 0.540 & \textbf{0.766} & 0.817 & 0.748\\
TABE$\mathsection$ & ResNet-101 & 0.746 & 0.541 & 0.617 & 0.809 & 0.808 \\
CLGR$\mathsection$ & ResNet-101 & 0.795 & \textbf{0.615} & 0.723 & 0.870 & 0.739\B\\
\hline
Baseline & ResNeXt-101 & \underline{\textbf{0.819}} & 0.616 & 0.768 & 0.853 & 0.744\T\\
LNTL$\mathsection$ & ResNeXt-101 & 0.776 & 0.597 & 0.746 & 0.821 & 0.778 \\
TABE$\mathsection$ & ResNeXt-101 & 0.817 & \underline{\textbf{0.674}} & \textbf{0.857} & \underline{\textbf{0.908}} & 0.768 \\
CLGR$\mathsection$ & ResNeXt-101 & 0.784 & 0.650 & 0.785 & 0.818 & \textbf{0.807}\B\\
\hline
Baseline & DenseNet & 0.775 & 0.559 & 0.655 & 0.851 & 0.695\T\\
LNTL$\mathsection$ & DenseNet & 0.760 & 0.548 & 0.750 & 0.859 & 0.689 \\
TABE$\mathsection$ & DenseNet & \textbf{0.809} & \textbf{0.622} & 0.743 & \textbf{0.863} & \textbf{0.788} \\
CLGR$\mathsection$ & DenseNet & 0.760 & 0.596 & \underline{\textbf{0.872}} & 0.843 & 0.776\B\\
\hline
Baseline & Inception-v3 & 0.762 & 0.528 & 0.671 & 0.784 & 0.605\T\\
LNTL$\mathsection$ & Inception-v3 & \textbf{0.784} & 0.556 & 0.729 & 0.809 & 0.583 \\
TABE$\mathsection$ & Inception-v3 & 0.751 & \textbf{0.593} & 0.735 & 0.818 & \textbf{0.746} \\
CLGR$\mathsection$ & Inception-v3 & 0.722 & 0.537 & \textbf{0.775} & \textbf{0.847} & 0.706\B\\

\hline
\hline
\end{tabu}
}
}
\label{tab:Instrument}
\end{table*}

\Cref{tab:Instrument} compares the generalisation ability of each instrument debiasing method against the baseline. We test the models on a number of datasets of differing distributions to test generalisation. We apply the debiasing heads to several different model architectures (EfficientNet-B3 \cite{tanEfficientNetRethinkingModel2019}, ResNet-101 \cite{heDeepResidualLearning2016}, ResNeXt-101 \cite{xieAggregatedResidualTransformations2017}, DenseNet \cite{huangDenselyConnectedConvolutional2017}, Inception-v3 \cite{szegedyRethinkingInceptionArchitecture2016a}) and compare the results, allowing us to select a champion architecture for further experimentation. ResNeXt-101 is chosen for focused experimentation and discussion since it achieves the highest score on 3 out of the 5 test sets, as seen in \Cref{tab:Instrument}. TABE and CLGR outperform the respective baselines across most experiments for most tested architectures, but all methods generally under-perform when used with ResNet-101 or when tested on the Atlas Dermoscopic data (please see \Cref{tab:Instrument} for detailed breakdown). On the MClass clinical test set using ResNeXt-101, the CLGR head is the difference between the model performing below the dermatologist benchmark, and exceeding it (8.5\% AUC increase), highlighting the potential impact of these domain generalisation methods.

In general, the greatest performance increases are observed on the clinical test sets, likely since these have the greatest domain shift from the dermoscopic training set. The models utilising LNTL were less successful, and even negatively impacted performance in some cases. This highlights that a single technique should not be applied in blanket fashion, as is done by \citet{bissotoDebiasingSkinLesion2020}; certain techniques may only be suitable for specific tasks and datasets.

\begin{figure}[htbp]
\includegraphics[width=\columnwidth]{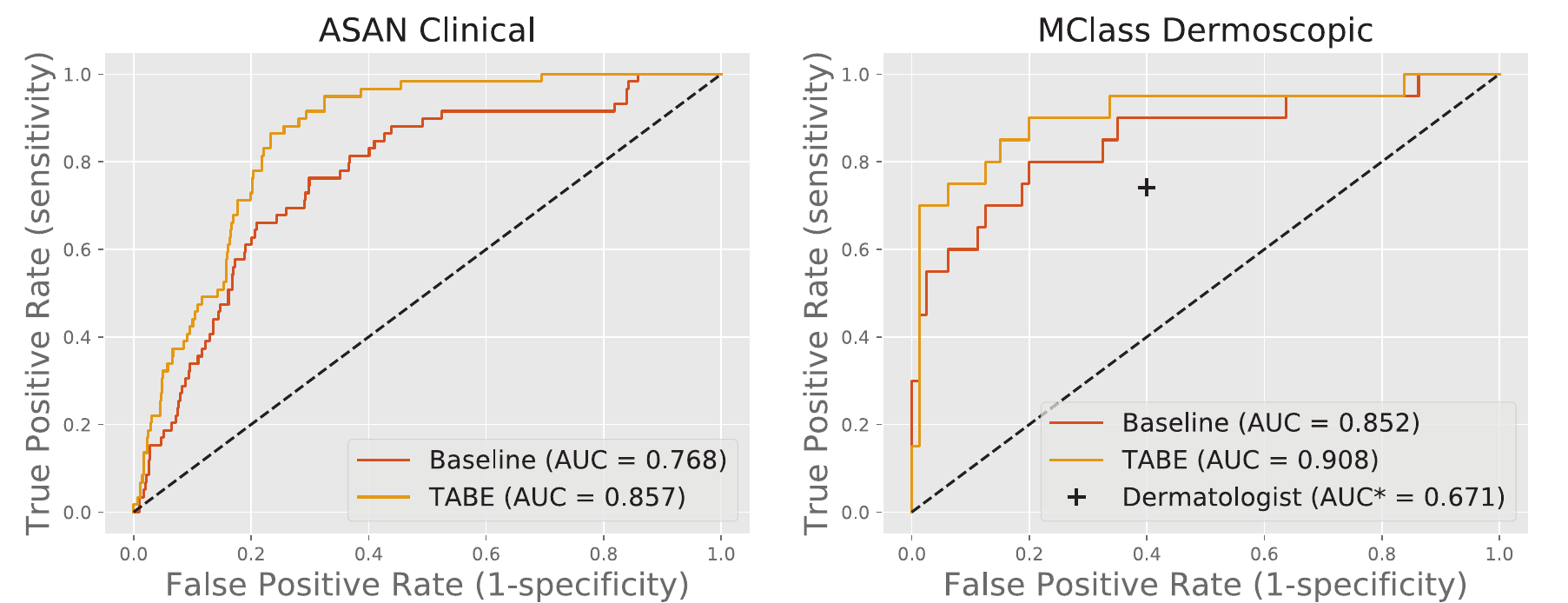}
\centering
 \caption{ROC curves for TABE instrument debiasing on ASAN clinical (left), and MClass dermoscopic (right) test sets, with ResNeXt-101 as the base architecture.}
\centering
\label{fig:InstrumClinic}
\end{figure}

\Cref{fig:InstrumClinic} shows the benefits of using a TABE head for instrument bias removal compared to the baseline model (both ResNeXt-101), showing an 11.6\% AUC improvement on the Asan clinical test set and a 6.6\% increase on the MClass dermoscopic test set. TABE can be differentiated from the baseline on each clinical test set, suggesting this to be a good tool for domain generalisation with dermoscopic and clinical data. Since both the training data and the MClass dermoscopic data are from the ISIC archive, the improved performance suggests the benefits of instrument invariance even when testing on data drawn from a similar distribution as the training data. This is likely due to the mitigation of domain bias caused by variation in the specific type of dermoscopic instrument used.

We experiment with using two debiasing heads, each removing a different bias (instrument, surgical marking or ruler), aiming to improve generalisation. The top configurations are shown in \Cref{tab:InstrumentDub}. Using a single TABE head removing instrument bias is still the most effective overall configuration. 
For results across a more complete set of configurations, refer to Section \ref{tab:InstrumentDubFull} in the Appendix.

\begin{table}[htbp]
\captionsetup[table]{skip=7pt}
\captionof{table}{Generalisation of ResNeXt-101 models trained using ISIC 2017 and 2020 data. The \enquote{dermatologists} row is the AUC scores from the work of \citet{brinkerComparingArtificialIntelligence2019}. Instrument, surgical marking and ruler labels represented by $\mathsection$, $\dagger$ and $\ddagger$ respectively.}
	\centering
	\resizebox{\linewidth}{!}{
		{\tabulinesep=0mm
			\begin{tabu}{@{\extracolsep{4pt}}c c c c c c@{}}
				\hline\hline
				\multicolumn{1}{c}{\multirow{2}{*}{Experiment}} & 
				\multicolumn{2}{c}{Atlas} &
				\multicolumn{1}{c}{Asan} &
				\multicolumn{2}{c}{MClass}\T\B \\
				\cline{2-3} \cline{4-4} \cline{5-6}
						& Dermoscopic & Clinical & Clinical & Dermoscopic & Clinical\T\B\\
				\hline\hline
Dermatologists & --- & --- & --- & 0.671 & 0.769\T\B\\
\hline

Baseline & 0.757 & 0.565 & 0.477 & 0.786 & 0.775\T\\

TABE$\mathsection$ & 0.817 & \textbf{0.674} & \textbf{0.857} & \textbf{0.908} & 0.768 \\

CLGR$\ddagger$ & 0.818 & 0.610 & 0.760 & 0.886 & \textbf{0.882} \\

TABE$\mathsection$+LNTL$\ddagger$ & \textbf{0.828} & 0.640 & 0.747 & 0.880 & 0.824\B\\
\hline
\hline
\end{tabu}
}
}
\label{tab:InstrumentDub}
\end{table}

\subsection{Ablation studies}

Ablation was built into the experimentation process since individual bias removal heads were implemented in isolation before attempting combinations, and debiasing heads were implemented both with and without gradient reversal. Using a single head to unlearn instrument bias is found to be more effective for generalisation than combining this head with artefact bias removal heads. TABE \cite{alviTurningBlindEye2019} both with and without the gradient reversal layer has proven successful for different tasks (\Cref{tab:RulerMarked}, \Cref{tab:Instrument}), but ablation of the gradient reversal layer from LNTL \cite{kimLearningNotLearn2019a} generally diminished performance (see \Cref{tab:Ablation}).

\begin{table}
\captionsetup[table]{skip=7pt}
\captionof{table}{Ablation of gradient reversal (denoted by *) from LNTL using ResNeXt-101 for removal of instrument bias.}
	\centering
	\resizebox{\linewidth}{!}{
		{\tabulinesep=0mm
			\begin{tabu}{@{\extracolsep{4pt}}c c c c c c@{}}
				\hline\hline
				\multicolumn{1}{c}{\multirow{2}{*}{Experiment}} & 
				\multicolumn{2}{c}{Atlas} &
				\multicolumn{1}{c}{Asan} &
				\multicolumn{2}{c}{MClass}\T\B \\
				\cline{2-3} \cline{4-4} \cline{5-6}
						& Dermoscopic & Clinical & Clinical & Dermoscopic & Clinical\T\B\\
				\hline\hline
LNTL & \textbf{0.804} & \textbf{0.612} & \textbf{0.768} & 0.819 & \textbf{0.801}\T\\

LNTL* & 0.783 & 0.605 & 0.710 & \textbf{0.827} & 0.747\B\\
\hline
\hline
\end{tabu}
}
}
\label{tab:Ablation}
\end{table}

\section{Limitations and future work}
\label{sec:limitations}

While we have demonstrated the impressive performance of unlearning techniques for artefact debiasing, one drawback of such approaches is the need to manually label these artefacts in each training image. These artefacts, however, are often quick and easy to identify by untrained individuals. Further research may look to uncover biases caused by other artefacts in a similar manner to the work of \citet{winklerAssociationDifferentScale2021,winklerAssociationSurgicalSkin2019} and evaluate the effectiveness of unlearning techniques at mitigating these. Future work could also incorporate an algorithm which accurately labels artefacts and dynamically changes the model architecture to apply the required bias removal heads for the task.

As for potential improvements in domain generalisation, image size cannot be universally assumed as a proxy for the imaging instrument across all datasets so we recommend the actual instrument model be recorded as metadata when collecting training data for melanoma classification. Further research could include incorporating an instrument identification system which could provide labels for the image acquisition instrument. Further work may also evaluate the effectiveness of debiasing techniques in improving generalisation for diagnostic smartphone apps such as those reviewed by \citet{ratUseSmartphonesEarly2018}. Detailed analysis into the specific reasons why certain debiasing methods seem to work better for certain tasks could also be a useful avenue for future research.

\section{Conclusion}
\label{sec:conclusions}
This work has compared and demonstrated the effectiveness of debiasing methods in the context of skin lesion classification. We have successfully automated the mitigation of the surgical marking and ruler bias presented by \citet{winklerAssociationDifferentScale2021,winklerAssociationSurgicalSkin2019} using unlearning techniques (\Cref{subsec:experiments:results:artefacts}). We have investigated the use of bias removal models on two test sets for each artefact bias, one comprising lesion images with no artefacts present and one comprising the same lesion images with artefacts present. We have shown that the debiasing models perform on par with the baseline on the images without artefacts, and better on the images with artefacts. Utilising confusion loss with gradient reversal for bias removal \cite{kimLearningNotLearn2019a} improves the baseline AUC by 15.3\% on lesion images with rulers present (\Cref{tab:RulerMarked}) whilst using LNTL \cite{kimLearningNotLearn2019a} improves performance by 6.1\% when surgical markings are present. This suggests the addition of these debiasing heads leads to a model more robust to each artefact bias, without compromising performance when no bias is present. Utilising these techniques could be an alternative to the behaviour change amongst dermatologists suggested by \citet{winklerAssociationDifferentScale2021,winklerAssociationSurgicalSkin2019}.

We have also provided evidence of the generalisation benefits of using unlearning techniques to remove instrument-identifying information from the feature representation of CNNs trained for the classification of melanoma (\Cref{subsec:experiments:results:domgen}). We have demonstrated this using the ISIC training data \cite{codellaSkinLesionAnalysis2018,rotembergPatientcentricDatasetImages2021}, with image size as a proxy for the imaging instrument. To test the generalisation capabilities of our bias removal approaches, we have used five popular skin lesion test sets with varying degrees of domain shift. Utilising the \enquote{Turning a Blind Eye} \cite{alviTurningBlindEye2019} debiasing head is generally the most effective method, most notably inducing an 11.6\% AUC increase compared to the baseline on the Asan dataset \cite{hanClassificationClinicalImages2018} when using a ResNeXt-101 feature extractor. Our models perform better than experienced dermatologists, consistently beating their average AUC score on the MClass test sets \cite{brinkerComparingArtificialIntelligence2019}. Generalisation methods such as this are powerful for ensuring consistent results across dermatology clinics, and may have utility in the emerging diagnostic app space \cite{ratUseSmartphonesEarly2018}, given that differences between smartphone cameras are likely to introduce spurious variation in a similar manner.

\section*{Acknowledgements}
We thank Holger Haensle \& team of Heidelberg University Department of Dermatology for providing the test dataset from \citet{winklerAssociationSurgicalSkin2019}; we thank Jochen Weber of ISIC for providing additional information on the ISIC dataset; we thank Chris Deotte for providing cropped and resized ISIC data.

\bibliographystyle{icml2022}
\bibliography{Skin-Deep-Unlearning.bib}

\clearpage

\appendix

\section*{\Large Appendix}

\normalsize
This section presents supplementary material that can be referenced to enhance the readers' understanding of the details of the work. We could not fit the general literature review for skin lesion classification in the paper, so this is presented in  Section \ref{appendix:adlit}. Samples of each training and test dataset are illustrated in Section \ref{appendix:Examples}, to give a feel for the images present in each. Justification for our choice of metrics is given in Section \ref{appendix:metrics}. Hyperparameter tuning results are presented in Section \ref{appendix:Val}. Additional experimental results in the form of ROC curves and tables that were not included in the main paper can be found in Section \ref{appendix:AdResults}. Our attempt at interpreting artefact bias using vanilla gradient saliency maps \cite{Simonyan14a} is  presented in Section \ref{subsubsec:adresults:artefact:saliency}.

\section{Skin lesion classification}
\label{appendix:adlit}

The task of classifying skin lesions using machine learning has received attention within the research community since as early as 1988, initially using traditional machine learning methods such as decision trees in combination with segmentation \cite{okurSurveyAutomatedMelanoma2018a}. Originally, lack of model sophistication, compute power and quality data meant that performance was not at the level of dermatologists. Like with many other areas of computer vision, the rise of convolutional neural networks and ever increasing compute power has seen the performance of skin lesion classification models rapidly increase to the point where there is evidence of machine learning techniques matching or even surpassing dermatologists at the task \cite{brinkerConvolutionalNeuralNetwork2019,haenssleManMachineDiagnostic2018a}. The power of deep learning to extract features has meant many modern models perform best without segmentation, and often use information in the surrounding skin in the classification task \cite{bissotoConstructingBiasSkin2019b}.

Skin diseases can be separated into many classes. On the most granular scale, skin diseases can be separated into neoplastic and non-neoplastic conditions. A neoplastic condition is an abnormal growth of cells known as a tumour, while a non-neoplastic skin condition refers to any other type of skin condition. We focus on neoplastic lesions in this work. These neoplastic lesions can be separated into benign (non-cancerous) and malignant (cancerous), which is a very important classification to make, since cancerous tissue has the ability to invade the rest of the body and ultimately cause fatality. On a more fine-grained level, lesions may be classified by specific disease, such as cyst, basal cell carcinoma or melanoma. In terms of classification in machine learning, it is possible to use specific diseases as classes for prediction, allowing malignancy to be also inferred from this classification \cite{haIdentifyingMelanomaImages2020a}. We opt for the more common binary approach of classifying using benign/malignant as classes.

\section{Examples of data}
\label{appendix:Examples}

\begin{figure}[htbp]
\includegraphics[width=\columnwidth]{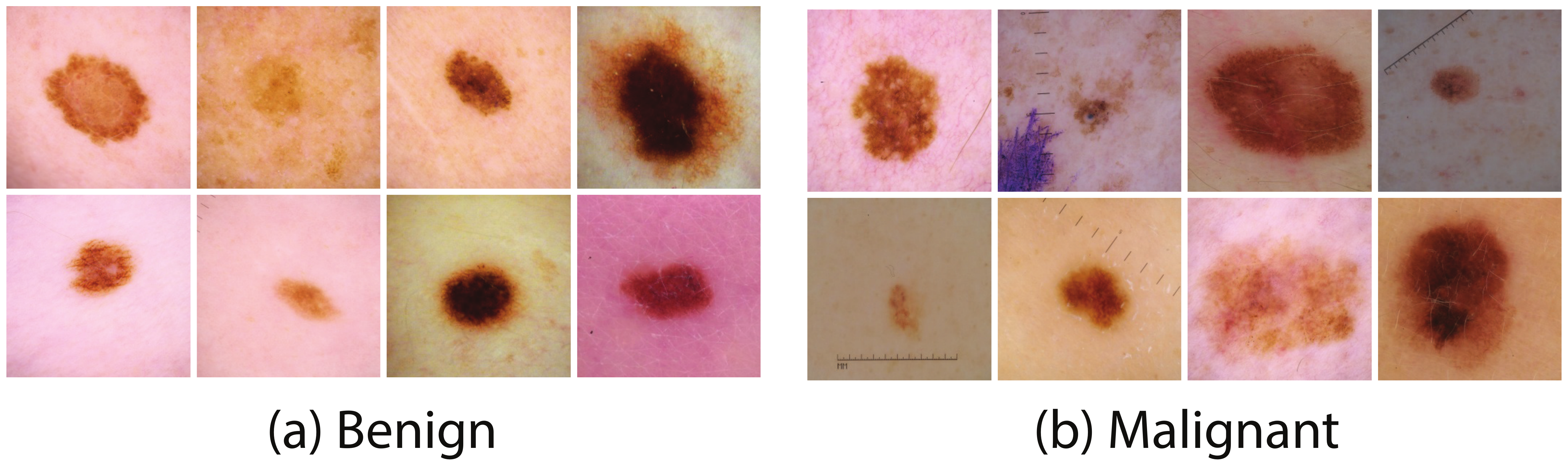}
\centering
  \caption{Example images from the ISIC dermoscopic training set \cite{codellaSkinLesionAnalysis2018,rotembergPatientcentricDatasetImages2021}.}
\centering
\label{fig:ISIC_ex}
\end{figure}

\Cref{fig:ISIC_ex} shows a sample of the images from the ISIC dermoscopic training data \cite{codellaSkinLesionAnalysis2018,rotembergPatientcentricDatasetImages2021}, including some examples of surgical markings and some examples of rulers.

\begin{figure}[htbp]
  \begin{subfigure}[b]{0.49\columnwidth}
\centering
    \includegraphics[width=\columnwidth]{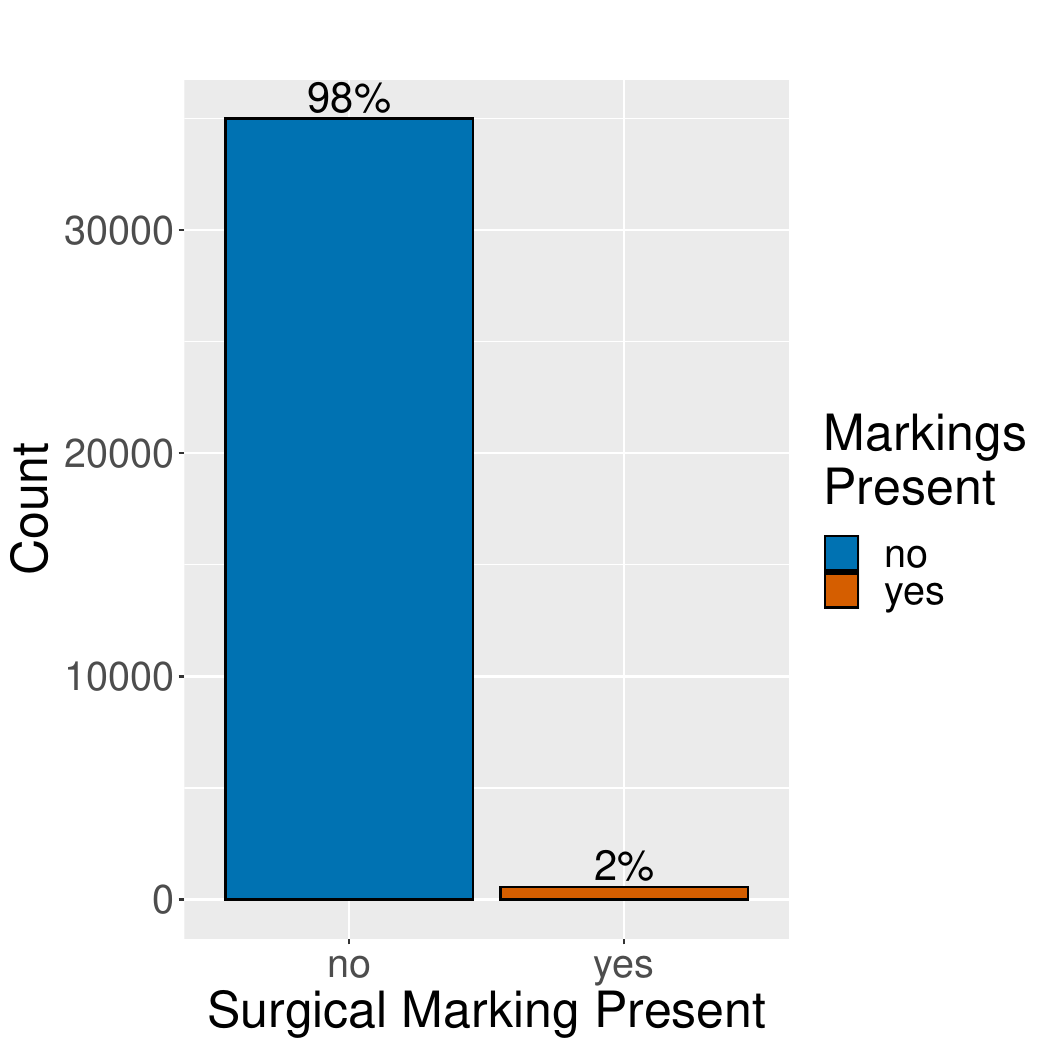}
    \caption{Marking distribution}
    \label{fig:M}
  \end{subfigure}
  \begin{subfigure}[b]{0.49\columnwidth}
\centering
    \includegraphics[width=\columnwidth]{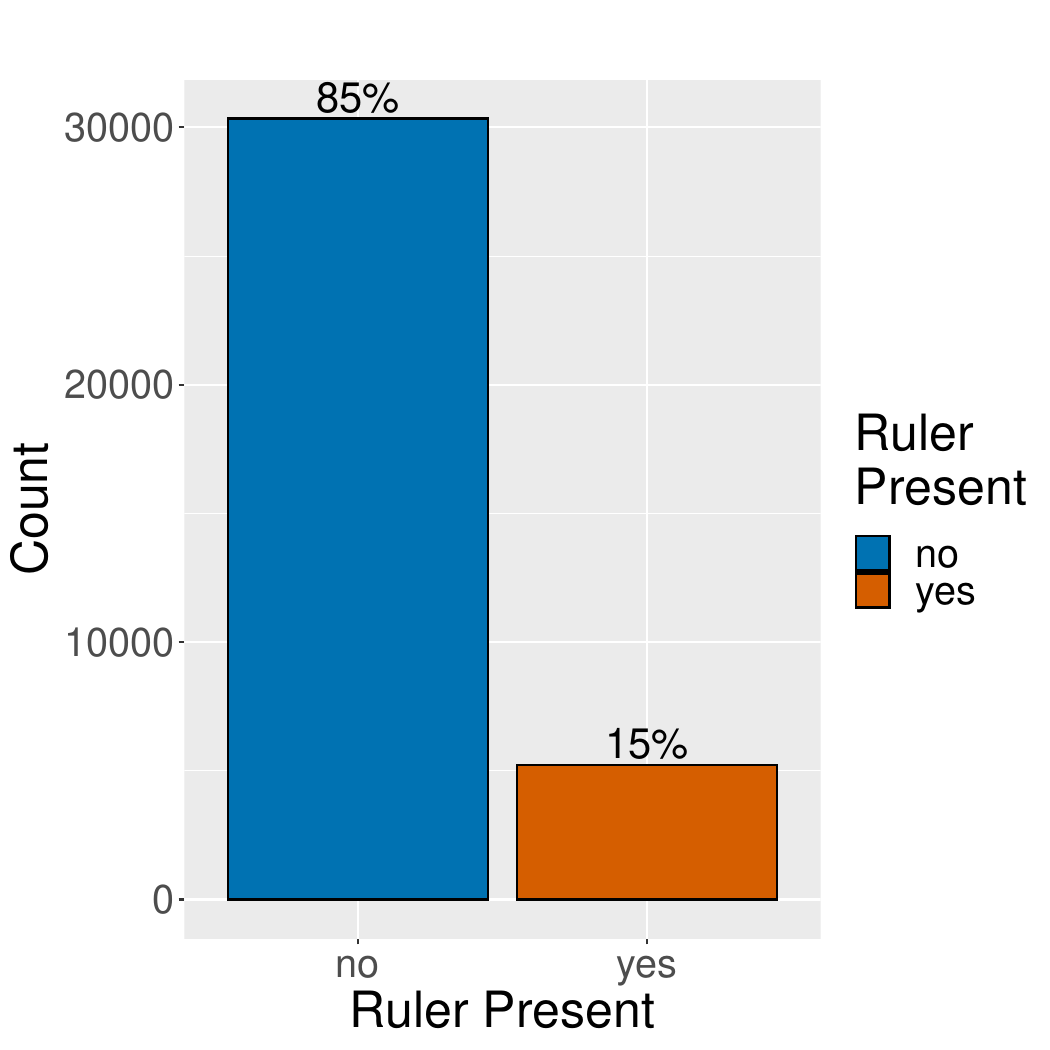}
    \caption{Ruler distribution}
    \label{fig:R}
  \end{subfigure}
  \caption{Class distribution of artefacts in ISIC 2020 \& 2017 training data \cite{codellaSkinLesionAnalysis2018,rotembergPatientcentricDatasetImages2021}.}
\centering
\label{fig:CDA}
\end{figure}

\Cref{fig:CDA} shows the class distribution for the surgical marking and ruler labels in the ISIC data. The distribution of artefacts is highly imbalanced, pointing to why weighted loss functions were needed to stabilise training.

\begin{figure}[htbp]
\includegraphics[width=\columnwidth]{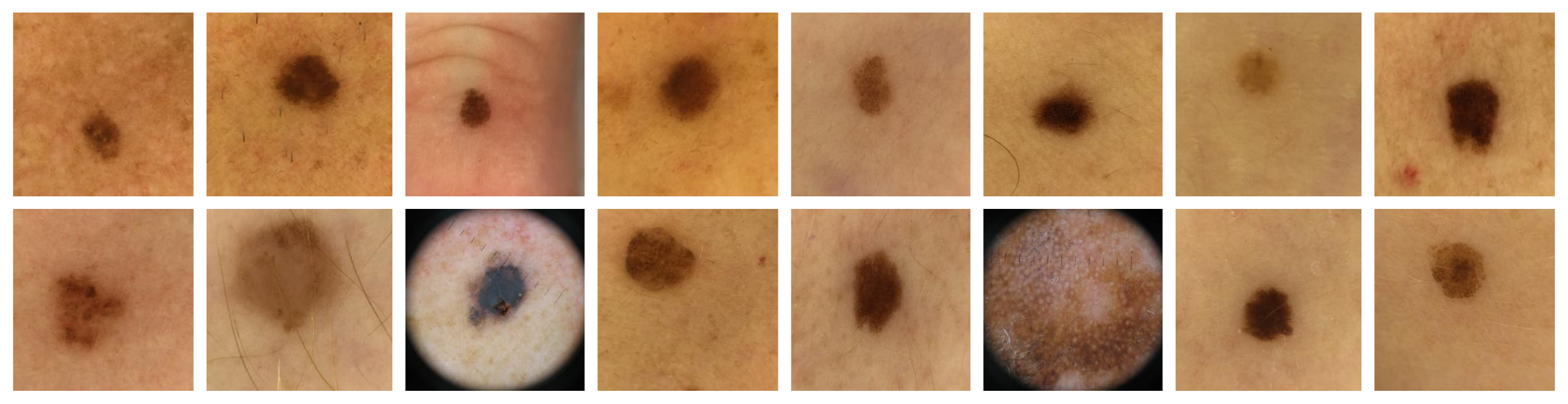}
\centering
\caption{Example images from the Heidelberg University training set with no artefacts \cite{winklerAssociationSurgicalSkin2019}.}
\label{fig:Heid_Plain_ex}
\end{figure}

\begin{figure}[htbp]
\includegraphics[width=\columnwidth]{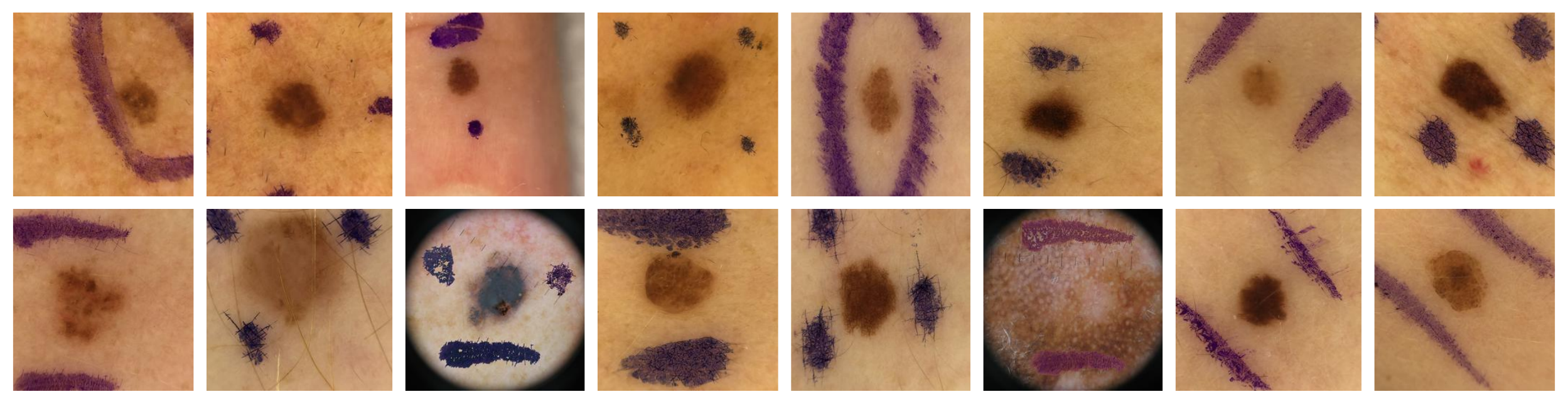}
\centering
\caption{Example images from the Heidelberg University training set with surgical markings \cite{winklerAssociationSurgicalSkin2019}.}
\label{fig:Heid_Marked_ex}
\end{figure}

\begin{figure}[htbp]
\includegraphics[width=\columnwidth]{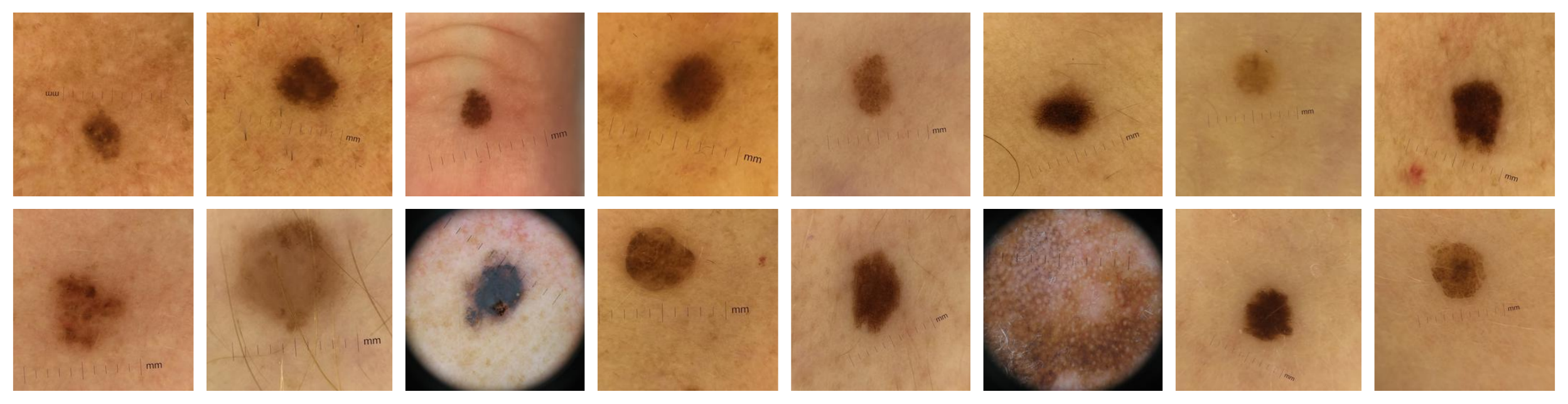}
\centering
\caption{Example images from the Heidelberg University training set with superimposed rulers \cite{winklerAssociationSurgicalSkin2019}.}
\label{fig:Heid_Ruler_ex}
\end{figure}

\Cref{fig:Heid_Plain_ex} shows a sample of the \enquote{Heid Plain} images from Heidelberg University \cite{winklerAssociationSurgicalSkin2019}. These are dermoscopic images collected by the university of a variety of neoplastic lesions. \Cref{fig:Heid_Marked_ex} shows a sample of the \enquote{Heid Marked} images from Heidelberg university \cite{winklerAssociationSurgicalSkin2019}. These are the same lesions from \enquote{Heid Plain}, but with surgical markings either applied in vivo (physically applied and images recaptured), or electronically superimposed. \Cref{fig:Heid_Ruler_ex} shows a sample of the \enquote{Heid Ruler} images, which was made by electronically superimposing rulers onto the \enquote{Heid Plain} images.

\begin{figure}[htbp]
\includegraphics[width=\columnwidth]{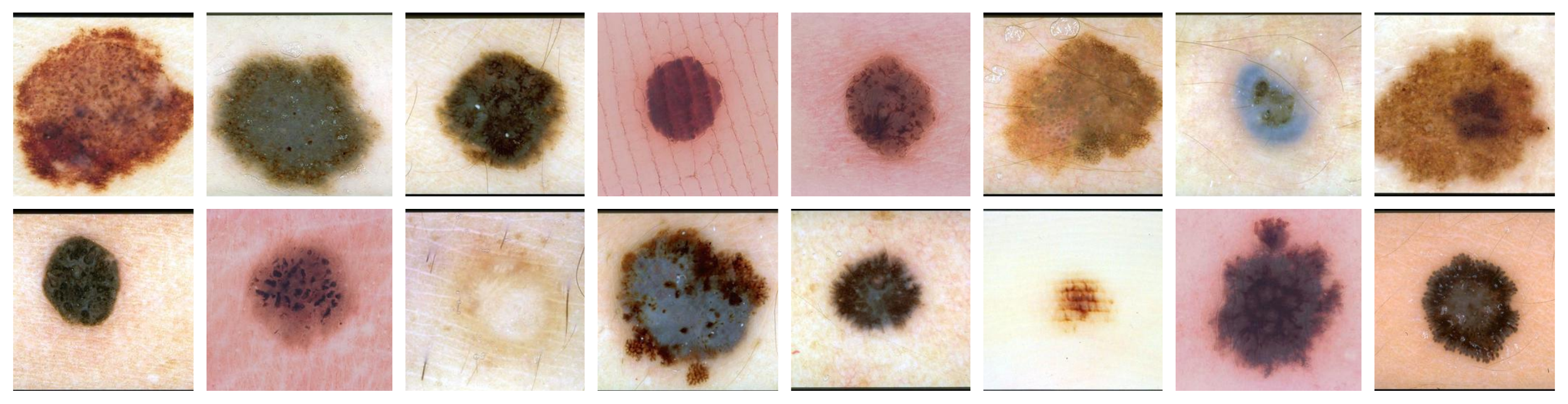}
\centering
\caption{Example images from the Interactive Atlas of Dermoscopy dermoscopic test set \cite{lioInteractiveAtlasDermoscopy2004}.}
\label{fig:AtlasD_ex}
\end{figure}

\begin{figure}[htbp]
\includegraphics[width=\columnwidth]{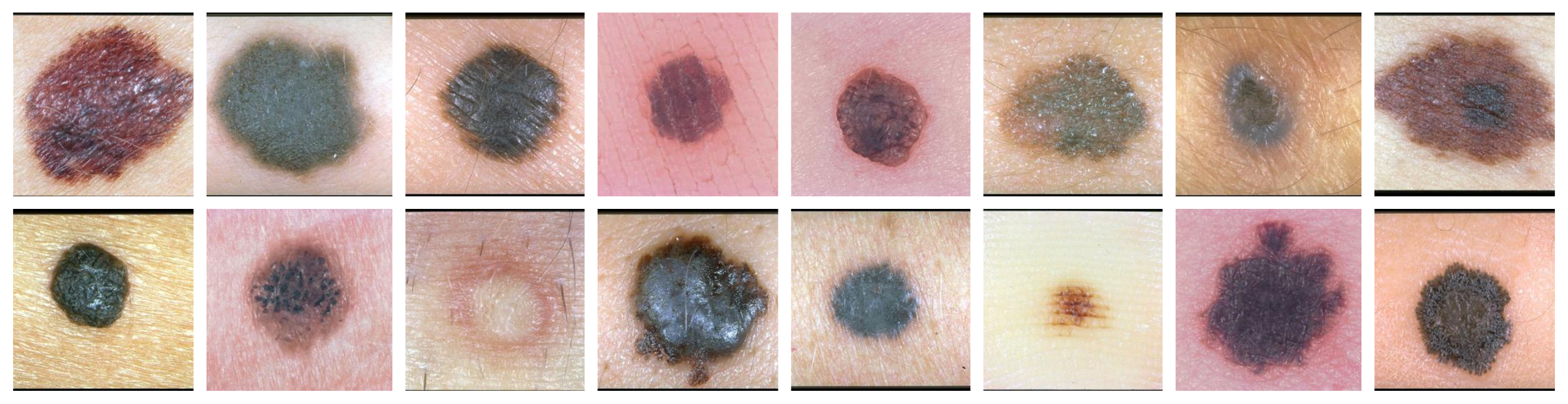}
\centering
\caption{Example images from the Interactive Atlas of Dermoscopy clinical test set \cite{lioInteractiveAtlasDermoscopy2004}.}
\label{fig:AtlasC_ex}
\end{figure}

\Cref{fig:AtlasD_ex} shows a sample of the \enquote{Interactive Atlas of Dermoscopy} \cite{lioInteractiveAtlasDermoscopy2004} \textit{dermoscopic} images, while \Cref{fig:AtlasC_ex} shows the equivalent \textit{clinical} images from the same set. The domain shift between clinical and dermoscopic images is clearly illustrated: the skin/lesion can be seen in more detail in the dermoscopic images due to the reduction in surface shine.

\begin{figure}[htbp]
\includegraphics[width=\columnwidth]{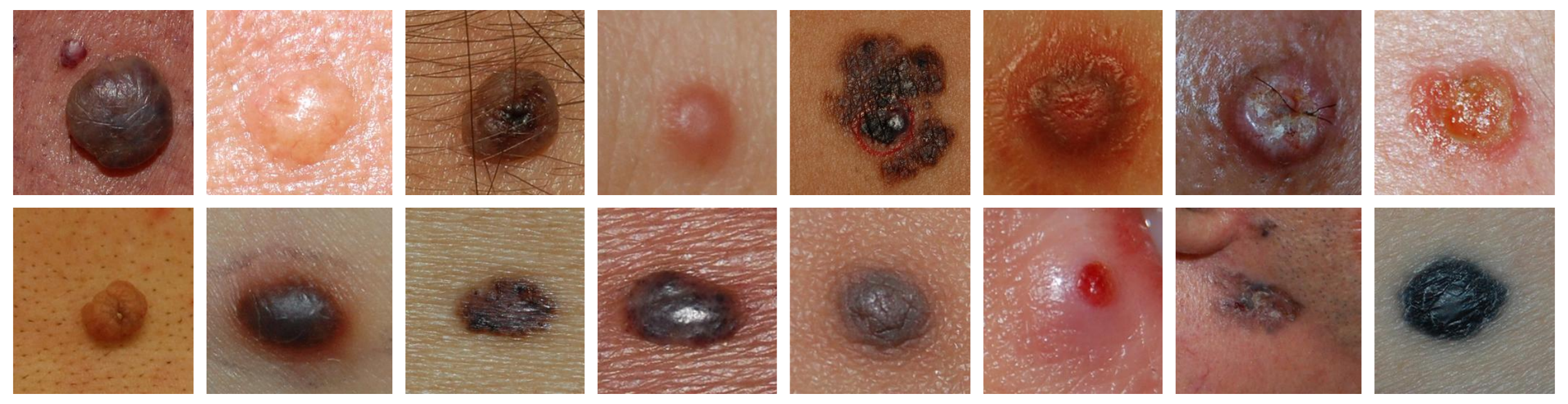}
\centering
\caption{Example images from the Asan clinical test set \cite{hanClassificationClinicalImages2018}.}
\label{fig:Asan_ex}
\end{figure}

\Cref{fig:Asan_ex} shows a sample of the Asan clinical test set \cite{hanClassificationClinicalImages2018}. This dataset is collected from the Asan medical centre, Seoul, South Korea and so features predominantly South Korean patients.

\begin{figure}[htbp]
\includegraphics[width=\columnwidth]{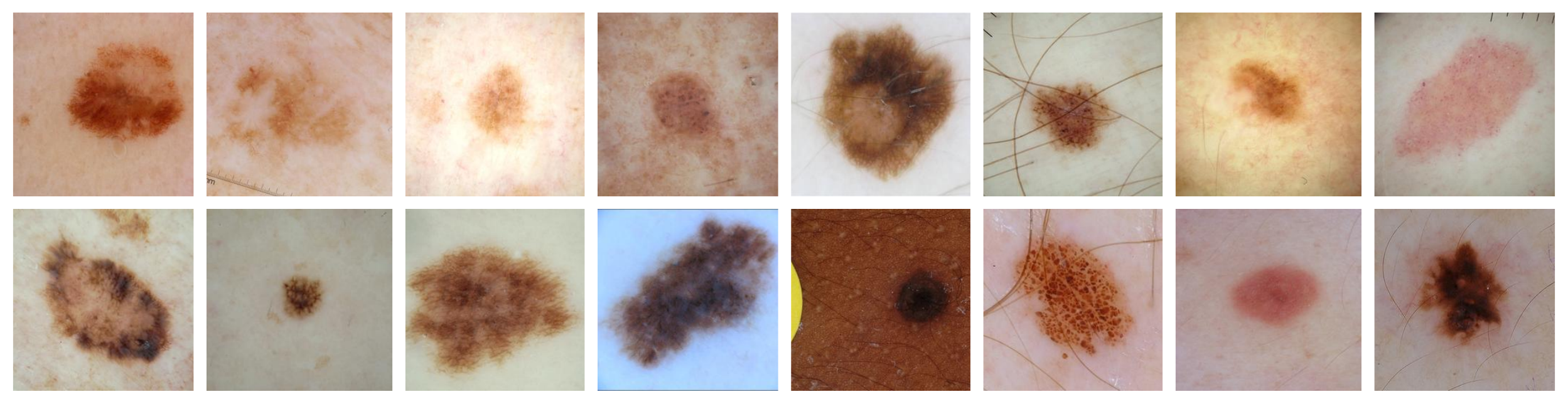}
\centering
\caption{Example images from the MClass dermoscopic test set \cite{brinkerComparingArtificialIntelligence2019}.}
\label{fig:MClassD_ex}
\end{figure}

\begin{figure}[htbp]
\includegraphics[width=\columnwidth]{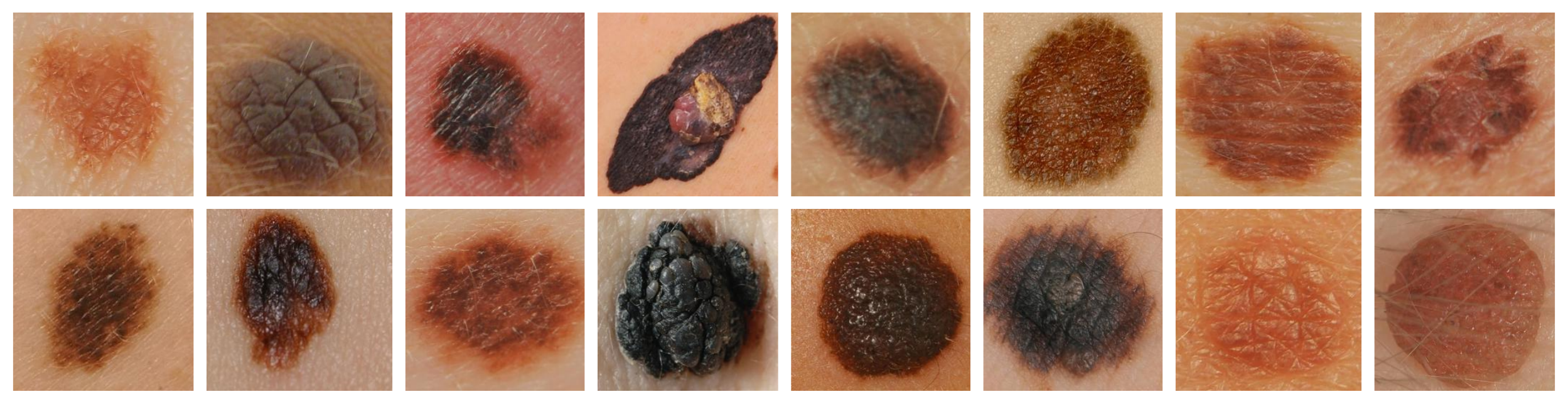}
\centering
\caption{Example images from the MClass clinical test set \cite{brinkerComparingArtificialIntelligence2019}.}
\label{fig:MClassC_ex}
\end{figure}

\Cref{fig:MClassD_ex} shows a sample of the MClass \cite{brinkerComparingArtificialIntelligence2019} dermoscopic benchmark test set, and \Cref{fig:MClassC_ex} shows a sample the MClass clinical benchmark test set. Both of these test sets were sent to a number of experienced dermatologists (157 for dermosocpic images, 145 for clinical images), who attempted to classify the images, with AUC scores reported in the work of \citet{brinkerComparingArtificialIntelligence2019}. Since true AUC cannot be calculated for dichotomous human predictions (we cannot adjust the threshold of human predictions), the authors use the average of sensitivity and specificity as a pseudo AUC score.

\section{Metrics}
\label{appendix:metrics}

The best evaluation metric for the task was carefully considered. We took into account the commonly used metrics in similar studies, as well as the specific requirements of the experiments.

Sensitivity (recall) is a measure of the proportion of the positive class that was correctly classified. Specificity is the proportion of the negative class that was correctly identified. These two metrics are defined as:
\[\textstyle Sensitivity=\frac{True\:Positive}{True\:Positive + False\:Negative}\] \\
\[\textstyle Specificity=\frac{True\:Negative}{True\:Negative + False\:Positive}\]
These are regularly used as metrics in the medical sciences, since it is important to both identify disease (leading to correct treatment) and rule disease out (preventing unnecessary treatment). In order to use these metrics, a threshold must be set at which the output of a model (between 0 and 1) is taken as a positive or negative classification. The default position for this threshold is 0.5, but this threshold may also be adjusted towards finding an acceptable trade-off between true positive, true negative, false positive and false negative predictions. Analysing the receiver operating characteristic (ROC) curve is a very useful way of finding this threshold, as it visualises how sensitivity and specificity vary over every possible threshold (see \Cref{fig:instrument_full} for example ROC curves).

The area under this curve (AUC) can hence be used as a single robust metric to evaluate the performance of a model where sensitivity and specificity are important, and where the threshold is open to adjustment (such as melanoma classification) \cite{mandrekarReceiverOperatingCharacteristic2010}. We avoid relying on accuracy, sensitivity and specificity in this work, since these all rely on the assumption of a selected threshold, and instead use AUC as the primary metric, also plotting the ROC curves. This is standard practice in melanoma classification \cite{haIdentifyingMelanomaImages2020a,hanClassificationClinicalImages2018,liSkinLesionAnalysis2018,okurSurveyAutomatedMelanoma2018a}.

We anticipate that the binary classification threshold would then be selected by a medical professional to suit their desired level of sensitivity and specificity. An AUC of 1 means the classifier is distinguishing positive and negative classes perfectly, and an AUC of 0.5 equates to random chance. Anything less than 0.5 and there may be an issue with the model or data labelling, since the model is actively predicting the wrong classes; in fact, inverting the data labels in this case would result in an AUC of over 0.5. We also avoid relying on accuracy due to the imbalance between benign/malignant lesions in the test sets meaning accuracy is not as descriptive of performance as AUC. 

\section{Hyperparameter tuning}
\label{appendix:Val}

\Cref{tab:Epochs} shows the chosen number of epochs for training each architecture for each dataset. These were chosen as the point at which the AUC reached its maximum or plateaued.

\Cref{tab:hyp} shows results of the grid search used to select learning rate and momentum, searching between 0.03 and 0.00001 for learning rate and 0 to 0.9 for momentum. We tune the baseline ResNeXt-101 model and also use these hyperparameters for the debiasing models to maximise cross-comparability. Whilst we used 5-fold cross validation for choosing the number of epochs, this was not computationally feasible for the grid search, and so a random subset (33\%, 3326 images) of the 2018 \cite{codellaSkinLesionAnalysis2018} challenge data is used as the validation set for hyperparameter tuning. With more time and computational resources, we could have optimised the number of epochs at the same time as these hyperparameters. In hindsight, perhaps a random search rather than a brute force grid search would have allowed more exhaustive tuning within the computational limitations but it is important to note that the optimal performance is not the primary focus of this paper and as such, a detailed hyperparameter tuning procedure does not significantly contribute to the objectives of this paper.

\begin{table}[htbp]
\captionsetup[table]{skip=7pt}
\captionof{table}{Optimal number of epochs for training, selected through analysis of cross validation curves.}
	\centering
	\resizebox{0.8\linewidth}{!}{
		{\tabulinesep=0mm
			\begin{tabu}{@{\extracolsep{12pt}}c c c@{}}
				\hline\hline
				Training dataset & Architecture & Epochs\T\B\\
				\hline\hline
 ISIC & EfficientNet-B3 & 15\T\\
 ISIC & ResNet-101 & 6 \\
 ISIC & ResNeXt-101 & 4 \\
 ISIC & Inception-v3 & 5 \\
 ISIC & DenseNet & 6\B\\

\hline
\hline
        \end{tabu}
    }
}
\label{tab:Epochs}
\end{table}

\begin{table*}[!t]
\captionsetup[table]{skip=7pt}
\captionof{table}{Hyperparamter tuning of baseline ResNeXt-101 model, trained for 4 epochs and using a random subset (33\%, 3326 images) of the 2018 ISIC challenge data \cite{codellaSkinLesionAnalysis2018} is used as the validation set for hyperparameter tuning.}
	\centering
	\resizebox{\linewidth}{!}{
		{\tabulinesep=0mm
			\begin{tabu}{@{\extracolsep{12pt}}c c c@{}}
				\hline\hline
				LR & Mom & AUC\T\B\\
				\hline\hline
 0.03 & 0 & 0.807\T\\
 0.01 & 0 & 0.825 \\
 0.003 & 0 & 0.837 \\
 0.001 & 0 & 0.815 \\
 0.0003 & 0 & 0.783 \\
 0.0001 & 0 & 0.681 \\
 0.00003 & 0 & 0.469 \\
 0.00001 & 0 & 0.398\B\\

\hline
\hline
        \end{tabu}
        \quad
			\begin{tabu}{@{\extracolsep{12pt}}c c c@{}}
				\hline\hline
				LR & Mom & AUC\T\B\\
				\hline\hline
 0.03 & 0.3 & 0.824\T\\
 0.01 & 0.3 & 0.826 \\
 0.003 & 0.3 & 0.852 \\
 0.001 & 0.3 & 0.820 \\
 0.0003 & 0.3 & 0.798 \\
 0.0001 & 0.3 & 0.727 \\
 0.00003 & 0.3 & 0.524 \\
 0.00001 & 0.3 & 0.409\B\\

\hline
\hline
        \end{tabu}
        \quad
			\begin{tabu}{@{\extracolsep{12pt}}c c c@{}}
				\hline\hline
				LR & Mom & AUC\T\B\\
				\hline\hline
 0.03 & 0.6 & 0.800\T\\
 0.01 & 0.6 & 0.837 \\
 0.003 & 0.6 & 0.854 \\
 0.001 & 0.6 & 0.826 \\
 0.0003 & 0.6 & 0.809 \\
 0.0001 & 0.6 & 0.770 \\
 0.00003 & 0.6 & 0.627 \\
 0.0001 & 0.6 & 0.445\B\\

\hline
\hline
        \end{tabu}
        \quad
			\begin{tabu}{@{\extracolsep{12pt}}c c c@{}}
				\hline\hline
				LR & Mom & AUC\T\B\\
				\hline\hline
 0.03 & 0.9 & 0.789\T\\
 0.01 & 0.9 & 0.834 \\
 0.003 & 0.9 & 0.848 \\
 0.001 & 0.9 & 0.843 \\
 \textbf{0.0003} & \textbf{0.9} & \textbf{0.866} \\
 0.0001 & 0.9 & 0.814 \\
 0.00003 & 0.9 & 0.783 \\
 0.00001 & 0.9 & 0.681\B\\

\hline
\hline
        \end{tabu}
    }
}
\label{tab:hyp}
\end{table*}

\section{Additional results}
\label{appendix:AdResults}

\subsection{Artefact bias removal}
\label{subsec:adresults:artefact}

To label the artefacts in the training data, we attempt to use colour thresholding to automatically label both surgical markings and rulers. We set the script to separate the images into different directories for inspection. This method is somewhat successful for identifying surgical markings. However, by looking at the images labelled unmarked, we see that some are not picked up, and so we also go through and manually pick out the remainders. This method does not work well at all for labelling rulers, likely due to the fact that hairs have similar pixel values to rulers. As a result, we manually label each image for rulers. The manual labelling process is not difficult to the human eye since these artefacts are quite obvious and so this can be done quickly and accurately.

\begin{figure*}[t]
  \begin{subfigure}[b]{\columnwidth}
\centering
    \includegraphics[width=\columnwidth]{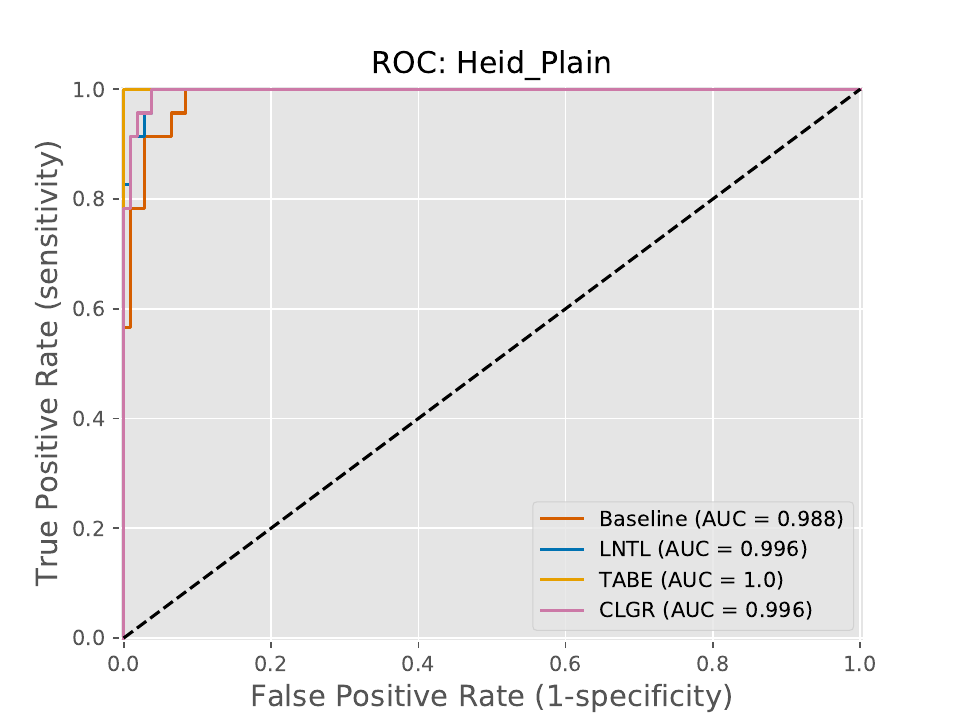}
    \caption{No surgical markings present.}
    \label{fig:HeidBlank1}
  \end{subfigure}
  \begin{subfigure}[b]{\columnwidth}
\centering
    \includegraphics[width=\columnwidth]{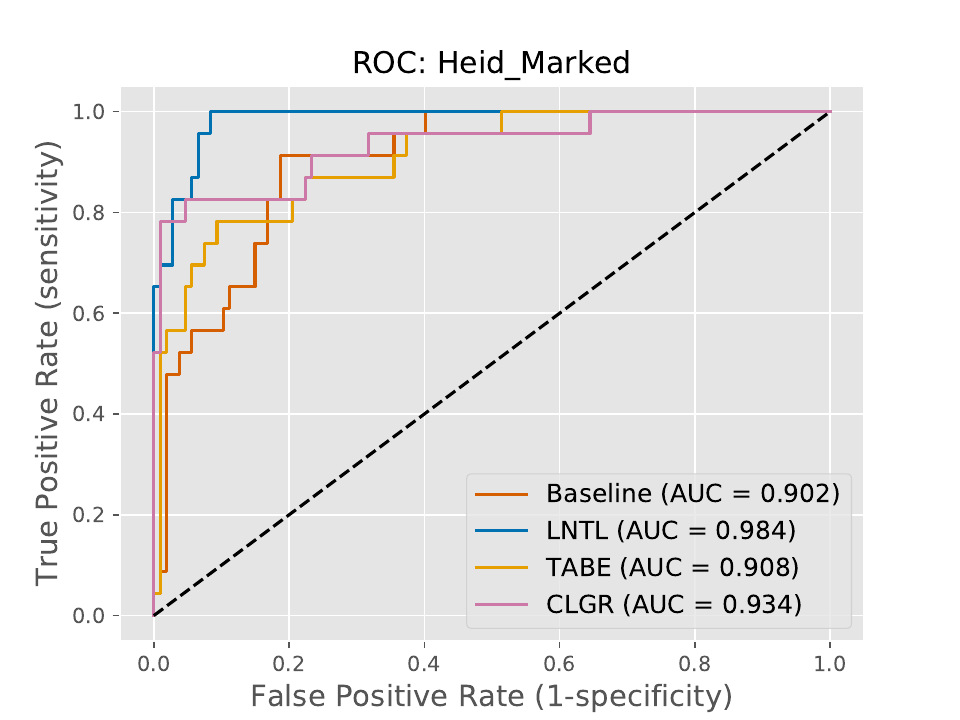}
    \caption{Surgical markings present.}
    \label{fig:HeidMarked}
  \end{subfigure}
  \caption{Comparison between model performances with no surgical markings present (left) vs with surgical markings present (right). \textbf{EfficientNet-B3} trained on ISIC 2020 \& 2017 data \cite{codellaSkinLesionAnalysis2018,rotembergPatientcentricDatasetImages2021}, skewed to $dm$=20.}
\centering
\label{fig:Heid_Marked}
\end{figure*}

\begin{figure*}[t]
  \begin{subfigure}[b]{\columnwidth}
\centering
    \includegraphics[width=\columnwidth]{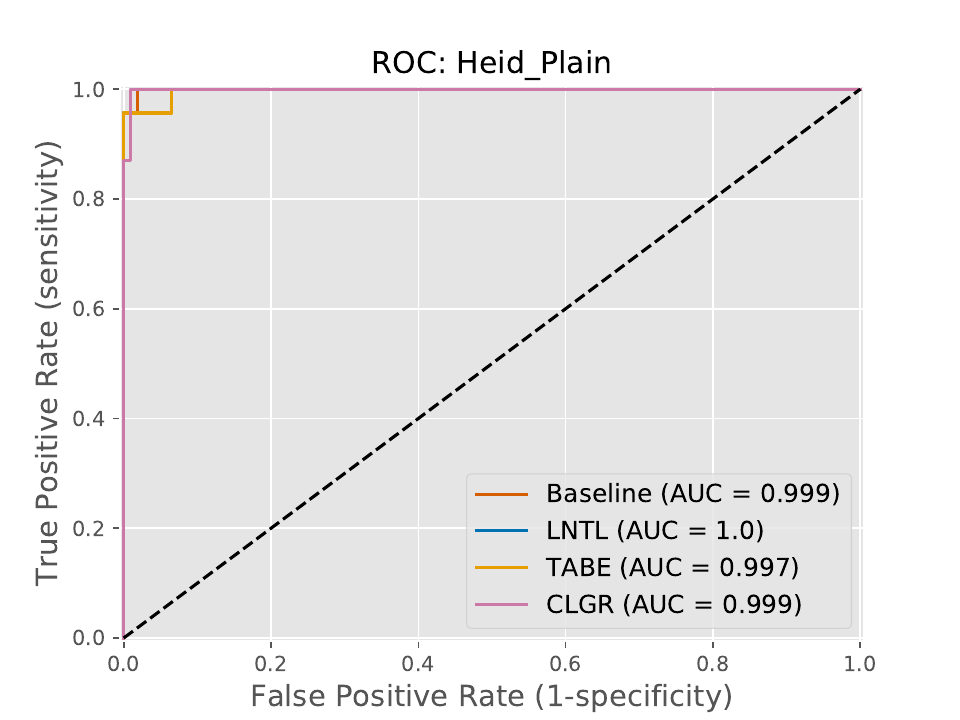}
    \caption{No rulers present.}
    \label{fig:HeidBlank2}
  \end{subfigure}
  \begin{subfigure}[b]{\columnwidth}
\centering
    \includegraphics[width=\columnwidth]{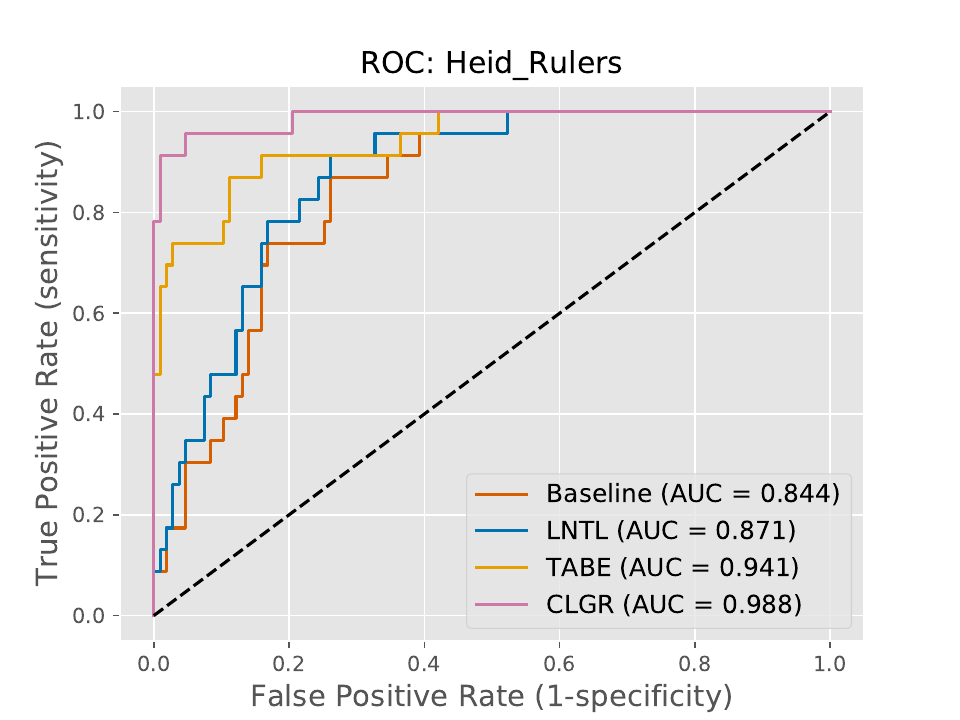}
    \caption{Rulers present.}
    \label{fig:HeidRuler}
  \end{subfigure}
  \caption{Comparison between model performances with no rulers present (left) vs with rulers present (right). \textbf{EfficientNet-B3} trained on ISIC 2020 \& 2017 data \cite{codellaSkinLesionAnalysis2018,rotembergPatientcentricDatasetImages2021}, skewed to $dr$=18.}
\centering
\label{fig:Heid_Ruler}
\end{figure*}

\Cref{fig:Heid_Marked} shows the ROC plots from the surgical marking bias experiments. All models perform almost perfect classification of the easy test set with no artefacts (\Cref{fig:HeidBlank1}). The test set with surgical markings present causes performance to drop for all models. However, it is clear from \Cref{fig:HeidMarked} that the debiasing models are more robust than the baseline, especially LNTL, which retains close to the same AUC score across both test sets. Similarly, the introduction of rulers into the lesion images also causes a drop in the performance of all models (see \Cref{fig:HeidRuler}). The baseline is again affected most by this bias, with TABE clearly most robust to it.

Although we choose to use AUC as the primary metric rather than accuracy since accuracy depends on the threshold set (qualified in Appendix \Cref{appendix:metrics}), \Cref{tab:RulerMarked_acc} shows the accuracy scores that correspond to the AUC scores in \Cref{tab:RulerMarked}. These accuracy scores are calculated with a threshold of 0.5. These accuracy scores also corroborate that the debiasing techniques improve the models robustness to artefact bias.

\begin{table*}[!t]
\captionsetup[table]{skip=7pt}
\captionof{table}{Comparison of each unlearning technique against the baseline, trained on artificially skewed ISIC data. \enquote{Heid plain} test set is free of artefacts while \enquote{Heid Marked} and \enquote{Heid Rulers} are the same lesions with surgical markings and rulers present respectively. All scores are \textbf{accuracy} (0.5 threshold).}
	\centering
	\resizebox{0.85\linewidth}{!}{
		{\tabulinesep=0mm
			\begin{tabu}{@{\extracolsep{12pt}}c c c@{}}
				\hline\hline
				\multicolumn{1}{c}{\multirow{2}{*}{Experiment}} & 
				\multicolumn{2}{c}{(a) Surgical Marking Removal ($dm$=20)}\T\B \\
				\cline{2-3}
						& Heid Plain & Heid Marked\T\B\\
				\hline\hline
Baseline & 0.903$\pm$0.006 & 0.853$\pm$0.018\T\\
LNTL$\dagger$ & 0.918$\pm$0.008 & 0.906$\pm$0.017\\
TABE$\dagger$ & \textbf{0.928}$\pm$0.023 & 0.836$\pm$0.058\\
CLGR$\dagger$ & 0.927$\pm$0.021 & \textbf{0.915}$\pm$0.015\B\\

\hline
\hline
        \end{tabu}
        \quad
			\begin{tabu}{@{\extracolsep{12pt}}c c c@{}}
				\hline\hline
				\multicolumn{1}{c}{\multirow{2}{*}{Experiment}} &
				\multicolumn{2}{c}{(b) Ruler Bias Removal ($dr$=18)}\T\B \\
				\cline{2-3}
						& Heid Plain & Heid Ruler\T\B\\
				\hline\hline
Baseline & 0.961$\pm$0.009 & 0.682$\pm$0.057\T\\
LNTL$\ddagger$ & 0.954$\pm$0.013 & 0.778$\pm$0.046\\
TABE$\ddagger$ & 0.955$\pm$0.009 & 0.835$\pm$0.030\\
CLGR$\ddagger$ & \textbf{0.963}$\pm$0.005 & \textbf{0.899}$\pm$0.002\B\\

\hline
\hline
        \end{tabu}
    }
}
\label{tab:RulerMarked_acc}
\end{table*}

\subsubsection{Saliency maps}
\label{subsubsec:adresults:artefact:saliency}

\begin{figure*}[htbp]
\includegraphics[width=\linewidth]{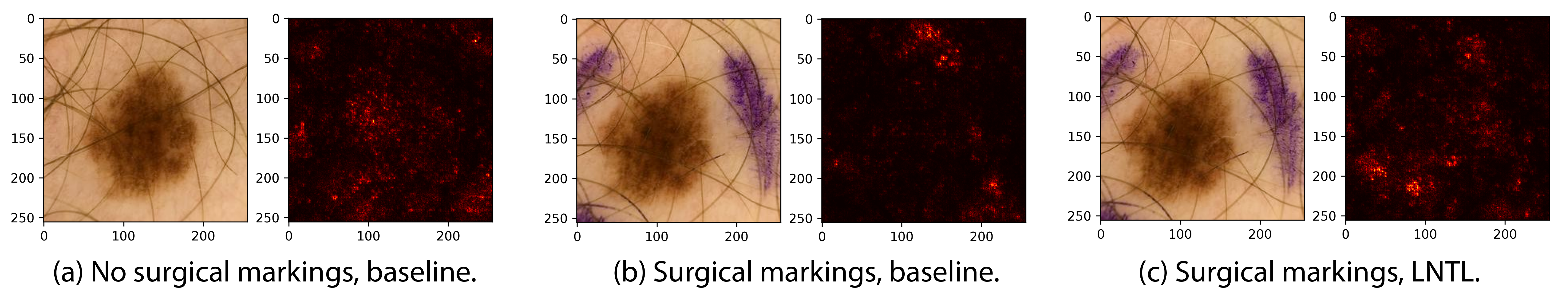}
\label{fig:HeidMarkedSaliencyLNTL}
\caption{Vanilla gradient saliency maps \cite{Simonyan14a} pointing to image regions most used by the model for classification. We compare the baseline on an unbiased and biased image of the same lesion, and also the LNTL model on the same biased image.}
\label{fig:saliencymap}
\end{figure*}

Since artefact bias can be located by image region, we attempt to identify whether the model is utilising the artefacts for classification by producing vanilla gradient saliency maps \cite{Simonyan14a}. This is a pixel attribution method and is designed to highlight pixels that were most relevant for classification using a heatmap of the same resolution as the input image. This method leverages backpropagation to calculate the gradient of the loss function with respect to the input pixels. These pixel-wise derivative values can then be used to create a heatmap of the input image which highlights the location of pixels with high values. We output saliency maps for both the plain and biased images from the work of \citet{winklerAssociationDifferentScale2021,winklerAssociationSurgicalSkin2019}, to see if the focus of the model shifts from the lesion to the artefact (see \Cref{fig:saliencymap}). 
It can be noticed in \Cref{fig:saliencymap} that for the baseline, there are less highlighted pixels in the lesion region when surgical markings are present compared to when there is not, and potentially more in regions that correspond to surgically marked regions. When using the LNTL model, the most salient pixels look to be located back in the general image region of the lesion, indicating the model has learned not to use the surgical markings for classification.

This method is a simple saliency map technique and suffers from certain issues, such as the ReLU activation function leading to a saturation problem \cite{shrikumarLearningImportantFeatures2019}. For future work, a more sophisticated technique like the GRAD-CAM post-hoc attention method \cite{selvarajuGradCAMVisualExplanations2017} may yield better quality visualisations.

\subsection{Domain generalisation}
\label{subsec:adresults:domgen}

\begin{figure*}[t]
  \begin{subfigure}[b]{0.45\textwidth}
\centering
    \includegraphics[width=\textwidth]{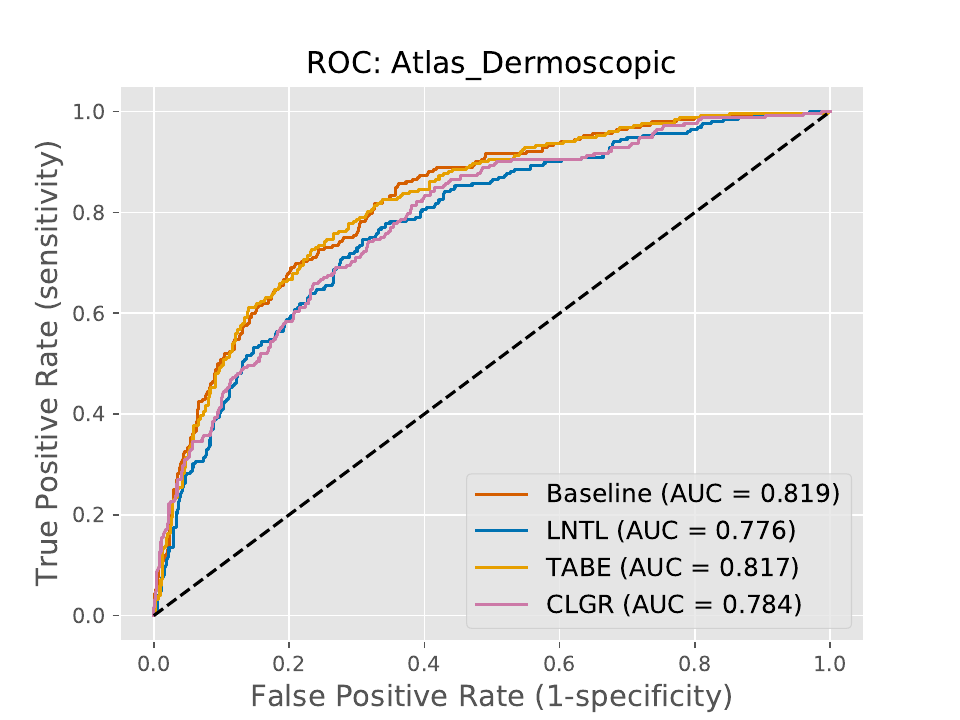}
    \caption{Atlas dermoscopic.}
    \label{fig:AtDerm}
  \end{subfigure}
  \begin{subfigure}[b]{0.45\textwidth}
\centering
    \includegraphics[width=\textwidth]{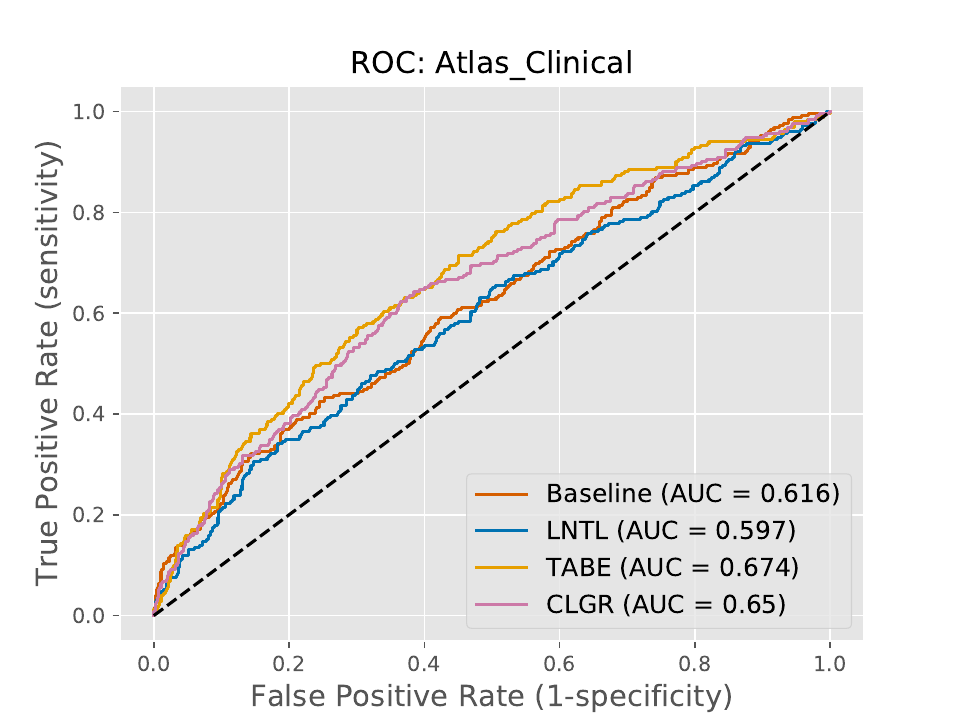}
    \caption{Atlas clinical}
    \label{fig:AtClin}
  \end{subfigure}
    \begin{subfigure}[b]{0.45\textwidth}
\centering
    \includegraphics[width=\textwidth]{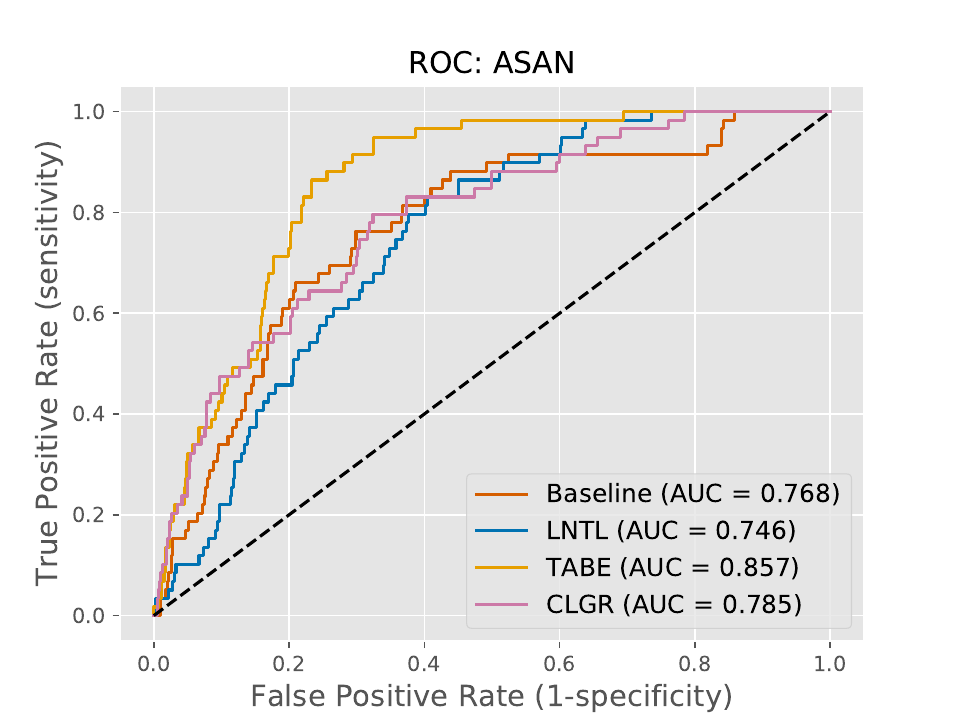}
    \caption{Asan (clinical)}
    \label{fig:AsanC}
  \end{subfigure}
    \begin{subfigure}[b]{0.45\textwidth}
\centering
    \includegraphics[width=\textwidth]{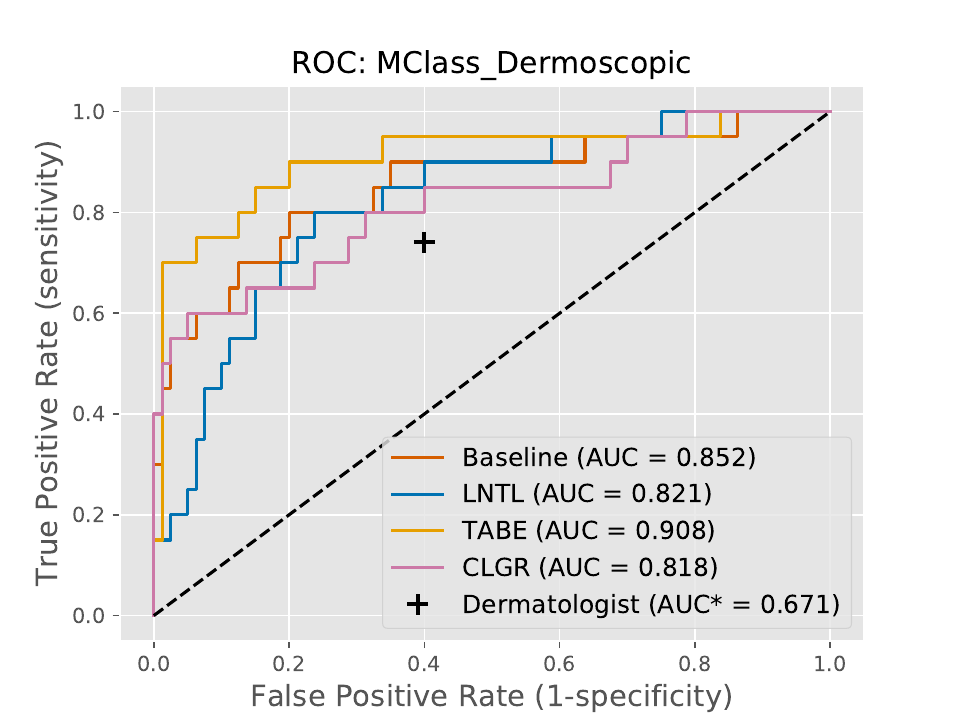}
    \caption{MClass Dermoscopic}
    \label{fig:MCLASSD}
  \end{subfigure}
    \begin{subfigure}[b]{0.45\textwidth}
\centering
    \includegraphics[width=\textwidth]{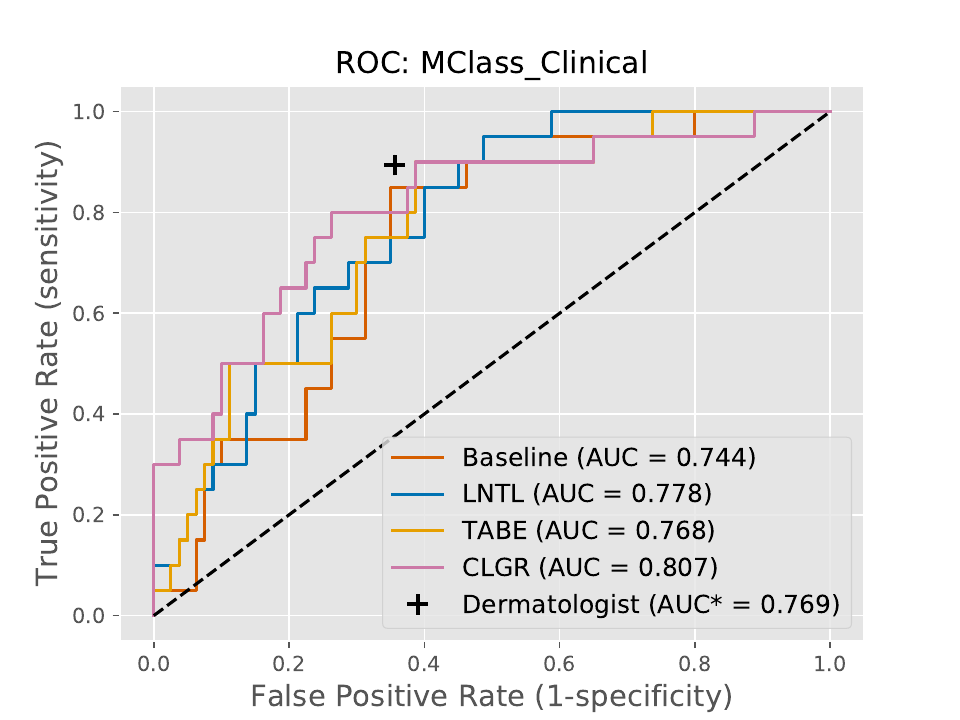}
    \caption{MClass Clinical}
    \label{fig:MCLASSC}
  \end{subfigure}
  \caption{ROC curves for each debiasing method, with ResNeXt-101 as the base architecture, aiming to remove spurious variation caused by the imaging instrument used. Model trained using the ISIC 2020 \cite{rotembergPatientcentricDatasetImages2021} and 2017 data \cite{codellaSkinLesionAnalysis2018} and tested on five test sets \cite{brinkerComparingArtificialIntelligence2019,hanClassificationClinicalImages2018,lioInteractiveAtlasDermoscopy2004}.}
\centering
\label{fig:instrument_full}
\end{figure*}

\Cref{fig:instrument_full} shows the ROC curves corresponding to \Cref{tab:Instrument}. TABE and CLGR are able to be differentiated from the baseline across most test sets, providing evidence that these models generalise better than the baseline when removing instrument bias.

\Cref{tab:InstrumentDubFull} is the full version of \Cref{tab:InstrumentDub}. A single debiasing head removing instrument bias is shown to be generally more effective than any combination of instrument, surgical marking or ruler bias removal. This is more evidence that combining debiasing heads can sometimes negatively impact performance, perhaps explaining the poor performance of the seven-head solution in the work of \citet{bissotoDebiasingSkinLesion2020}.

\begin{table*}[!t]
\captionsetup[table]{skip=7pt}
\captionof{table}{\textit{Domain generalisation}: Comparison of generalisation benefits of using different targets for the debiasing heads (ResNeXt-101), including some models with two debiasing heads. The \enquote{dermatologists} row is the AUC scores from the work of \citet{brinkerComparingArtificialIntelligence2019}. A capital \enquote{D} indicates the images are dermoscopic, while a capital \enquote{C} means the images are clinical. The $\mathsection$ symbol indicates the use of instrument labels, $\dagger$ represents surgical marking labels and $\ddagger$ represents ruler labels.}
	\centering
	\resizebox{0.65\linewidth}{!}{
		{\tabulinesep=0mm
			\begin{tabu}{@{\extracolsep{4pt}}c c c c c c@{}}
				\hline\hline
				\multicolumn{1}{c}{\multirow{2}{*}{Experiment}} & 
				\multicolumn{2}{c}{Atlas} &
				\multicolumn{1}{c}{Asan} &
				\multicolumn{2}{c}{MClass}\T\B \\
				\cline{2-3} \cline{4-4} \cline{5-6}
						& Dermoscopic & Clinical & Clinical & Dermoscopic & Clinical\T\B\\
				\hline\hline
 Dermatologists & --- & --- & --- & 0.671 & 0.769\T\B\\
 \hline
 Baseline & 0.819 & 0.616 & 0.768 & 0.853 & 0.744\T\B\\
 \hline
 LNTL$\mathsection$ & 0.776 & 0.597 & 0.746 & 0.821 & 0.778\T\\
 TABE$\mathsection$ & 0.817 & \textbf{0.674} & \textbf{0.857} & \textbf{0.908} & 0.768 \\
 CLGR$\mathsection$ & 0.784 & 0.650 & 0.785 & 0.818 & 0.807\B\\
 \hline
 LNTL$\dagger$ & 0.737 & 0.589 & 0.631 & 0.731 & 0.799\T\\
 TABE$\dagger$ & 0.788 & 0.658 & 0.768 & 0.889 & 0.851 \\
 CLGR$\dagger$ & 0.758 & 0.583 & 0.679 & 0.819 & 0.774\B\\
 \hline
 LNTL$\ddagger$ & 0.818 & 0.616 & 0.705 & 0.849 & 0.759\T\\
 TABE$\ddagger$ & 0.813 & 0.667 & 0.679 & 0.865 & 0.846 \\
 CLGR$\ddagger$ & 0.818 & 0.610 & 0.760 & 0.886 & \textbf{0.882}\B\\
 \hline
 LNTL$\mathsection$+LNTL$\dagger$ & 0.789 & 0.588 & 0.704 & 0.849 & 0.796\T\\
 TABE$\mathsection$+TABE$\dagger$ & 0.807 & 0.629 & 0.779 & 0.859 & 0.810 \\
 LNTL$\mathsection$+TABE$\dagger$ & 0.802 & 0.591 & 0.766 & 0.864 & 0.705 \\
 LNTL$\mathsection$+CLGR$\dagger$ & 0.573 & 0.574 & 0.645 & 0.717 & 0.617 \\
 CLGR$\mathsection$+CLGR$\dagger$ & 0.801 & 0.656 & 0.840 & 0.811 & 0.820 \\
 CLGR$\mathsection$+LNTL$\dagger$ & 0.763 & 0.615 & 0.767 & 0.833 & 0.790 \\
 TABE$\mathsection$+LNTL$\dagger$ & 0.823 & 0.629 & 0.787 & 0.881 & 0.781\B\\
 \hline
 LNTL$\mathsection$+LNTL$\ddagger$ & 0.786 & 0.604 & 0.686 & 0.837 & 0.779\T\\
 TABE$\mathsection$+TABE$\ddagger$ & 0.806 & 0.612 & 0.783 & 0.827 & 0.794 \\
 LNTL$\mathsection$+TABE$\ddagger$ & 0.806 & 0.606 & 0.728 & 0.881 & 0.747 \\
 LNTL$\mathsection$+CLGR$\ddagger$ & 0.816 & 0.618 & 0.740 & 0.872 & 0.792 \\
 CLGR$\mathsection$+CLGR$\ddagger$ & 0.798 & 0.613 & 0.723 & 0.898 & 0.795 \\
 CLGR$\mathsection$+LNTL$\ddagger$ & 0.793 & 0.586 & 0.704 & 0.876 & 0.776 \\
 TABE$\mathsection$+LNTL$\ddagger$ & \textbf{0.828} & 0.640 & 0.747 & 0.880 & 0.824\B\\

\hline
\hline
\end{tabu}
}
}
\label{tab:InstrumentDubFull}
\end{table*}

\end{document}